\documentclass{article}

\usepackage[main, final]{neurips_2026}

\usepackage[utf8]{inputenc} 
\usepackage[T1]{fontenc}    
\usepackage{hyperref}       
\usepackage{url}            
\usepackage{booktabs}       
\usepackage{amsfonts}       
\usepackage{nicefrac}       
\usepackage{microtype}      
\usepackage{xcolor}         
\usepackage{subcaption}
\usepackage{placeins} 
\usepackage{calc}
\usepackage{multirow}
\usepackage{graphicx}%
\usepackage{float}
\usepackage{xspace}
\usepackage{tcolorbox}
\usepackage{enumitem}
\usepackage{ulem}

\newcommand{\name}{CCDF\xspace}

\title{CCDF: A Benchmark Dataset for Deepfake Detection in Surveillance Footage}

\author{%
  Baptiste Chopin$^1$, Thomas Swearingen$^2$, Arun Ross$^2$, Antitza Dantcheva$^3$, Christian Rathgeb$^1$
    \\
  $^1$da/sec – Biometrics and Security Research Group, Hochschule Darmstadt     \\
  $^2$ Computer Science and Engineering Department, Michigan State University\\
  $^3$ STARS team, Inria Center at Université Côte d'Azur \\
}

\begin{document}

\maketitle
\begin{figure}[H]
  \centering
   \includegraphics[width=1.0\linewidth]{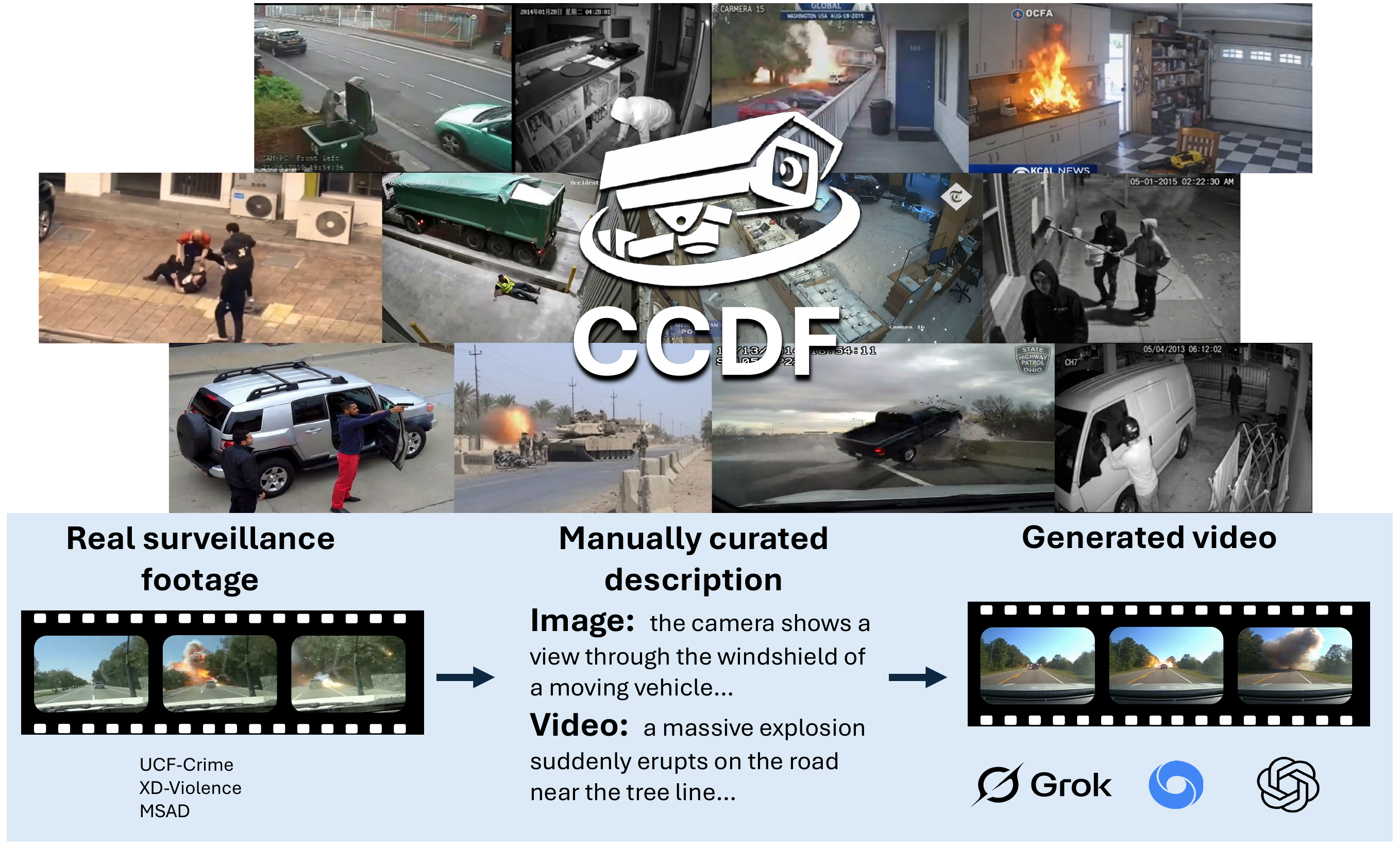}
   \label{fig:intro}
\end{figure}

\begin{abstract}
Due to rapid advances in Generative AI, commercial video generation tools can be used to produce fabricated surveillance footage that can fool both human viewers and automated synthetic video detectors.
Since these tools are so widely accessible, a malicious user can create a harmful video clip at minimal cost. The production and dissemination of such videos in high-stakes settings, such as crime reporting and elections, can misdirect emergency response efforts or distort political discourse.
Existing deepfake video datasets, used by the research community to develop deepfake detection algorithms, exhibit two limitations: (1)~they emphasize benign web content rather than footage of possibly malicious activity, and (2)~they rely on older or open-source generators that do not represent recent advances in generative systems.
We assemble CCtv DeepFakes (\name), a video deepfake dataset, to address both gaps.
\name contains 1{,}840 videos (460 real and 1{,}380 generated) spanning 16 crime and accident categories, with generated content produced using three leading commercial systems: Grok Imagine, Google VEO 3.1, and OpenAI Sora 2. \name is a highly realistic, small-scale, manually annotated dataset targeting evaluation of detection models.
We release three versions of the dataset: the \textit{raw} generated data, a \textit{cleaned} version in which resolution, frame rate, duration, and bitrate are standardized between real and synthetic samples to prevent detectors from exploiting trivial cues, and an \textit{altered} version simulating low-effort post-processing attacks.
We evaluate \name with ten recent state-of-the-art detectors covering different detection approaches.
Our results suggest that these approaches do not reliably distinguish CCDF’s generated videos from real ones, despite their strong reported performance on existing datasets.
These results further confirm the benefits of the CCDF dataset for advancing research in DeepFake detection.
Access to  \name can be requested by filling the form at \url{https://docs.google.com/forms/d/e/1FAIpQLScSmOnvIXIsAItzLosiUUpAWVnW7LeukjgODTeF4g3kv9eqJA/viewform}.
\end{abstract}

\section{Introduction}

The 2020s have seen a remarkable progression in the ability of \textit{generative AI} systems to produce synthetic text, audio, images, and videos.
The intention of such systems is to create societal benefits through lowering the cost of creating, understanding, and applying knowledge across domains like science, business, and education~\cite{Bommasani2021fmodels}.
This same capability, however, also lowers the cost of fabricating multimedia to a level almost anyone can afford as a video clip can be generated for little or no cost.
This is especially concerning in the case of synthetic video generation given the role visual media serves in confirming or refuting accounts of events~\cite{Rini2020dfsociety}.
The class of synthetic media most associated with this threat are \textit{deepfakes}\footnote{The other broad class are ``cheapfakes'' or ``shallowfakes'' which are manipulated media produced without deep learning, such as selective editing via photo editing software or mislabeling.}, defined by Mirsky and Lee~\cite{Mirsky2021dfsurvey} as ``believable media generated by a deep neural network.''
Early attempts to combat this problem were person-centric deepfake detectors, targeting the detection of synthetic faces and voices~\cite{Li2020xray, qian2020f3, Cozzolino2021idreveal} as the most salient harms were then perceived as impersonation, non-consensual imagery, and identity fraud.
As generative video systems developed, manufacturing or altering an \textit{event} such as a crime, an accident, or an act of political violence became more feasible.
The harm in this case is less to a specific individual and more to communities as a whole through the manipulation of public understanding via convincing footage of events that never occurred or occurred differently to what a video might suggest.

In response to this advancing capability, \textit{deepfake detection} has expanded from person-centric methods to more general-content detectors~\cite{interno2026aigenerated, park2026vidguardr,Kundu2025unite}, supported by new datasets containing arbitrary synthetic video~\cite{wang2024vidprom, Chen2024genvideo, Ma2026aigvd, liu2025lavid}.
However, current deepfake detection datasets have some limitations which make assessing detector performance in realistic conditions more difficult.
Specifically, they emphasize benign web content like animation, animal videos, or fantasy scenes that an attacker would be unlikely to fabricate, while under-representing more plausibly malicious content like scenes depicting arson, assault, or vandalism.
They also rely heavily on older or open-source generators whose outputs are often visibly synthetic and include only limited content from ``off-the-shelf'' commercial systems that an attacker would be more likely to use.
In addition, they exhibit video format mismatches between their real and generated subsets (\textit{e.g.,} differences in resolution, frame rate, duration, or bitrate) that allow detectors to exploit trivial cues rather than learning more robust signals.

Deviating from the above, we attempt to address these gaps with \textbf{CCtv DeepFakes} (\textbf{\name}), a video deepfake benchmark designed around a more realistic scenario: synthetic \textit{surveillance footage of crimes and accidents} generated by frontier commercial systems with video format properties standardized to prevent shortcut learning.
\name contains 1{,}840 videos (460 real and 1{,}380 generated) spanning 16 crime and accident categories, with synthetic content produced by three leading commercial video generators: xAI's Grok Imagine, Google's VEO 3.1, and OpenAI's Sora 2.
Real videos are sourced from established surveillance anomaly datasets~\cite{sultani2018real, Wu2020not, msad2024} with generation prompts derived from these real source videos via a vision-language model (which are also verified by human-reviewers to ensure correctness).
We provide three versions of this dataset: (1) the raw generated data (2) a \textit{cleaned} version of the dataset in which resolution, frame rate, and bitrate are standardized between real and generated samples, and (3) an \textit{altered} version simulating low-effort post-processing attacks.

We benchmark \name against ten recent state-of-the-art detectors spanning both, image-based and video-based approaches. \textbf{None of the ten detectors reliably distinguish \name's generated videos from real ones}, despite all reporting strong performance on existing benchmarks. 

To summarize, our contributions are as follows:
\begin{itemize}
\item We introduce \name, the first video deepfake benchmark targeting plausibly malicious content (surveillance footage of crimes and accidents) generated by frontier commercial systems (Sora 2, VEO 3.1, Grok Imagine), with standardized video encoding properties between the real and synthetic videos.
\item We construct \name through a vision-language-model-guided prompting pipeline that grounds synthetic content in real surveillance footage, and we release the raw data as well as a encoding-controlled \textit{cleaned} version and an \textit{altered} version simulating low-effort post-processing attacks.
\item We show that ten recent state-of-the-art deepfake detectors fail on \name, despite their strong reported performance on existing benchmarks. This exposes a gap between current evaluation datasets and the threat conditions detectors would likely face in deployment.
\item \name provides a small scale, highly realistic evaluation benchmark for deepfake detection models. Each video is generated from a prompt based on accurate and human verified descriptions of real event captured on CCTV. Unlike existing datasets, thanks to our \textit{cleaned} version, \name provides data with limited resolution and framerate biases that could be leveraged by detection methods.

\end{itemize}

\begin{figure}[t]
  \centering
   \includegraphics[width=1.0\linewidth]{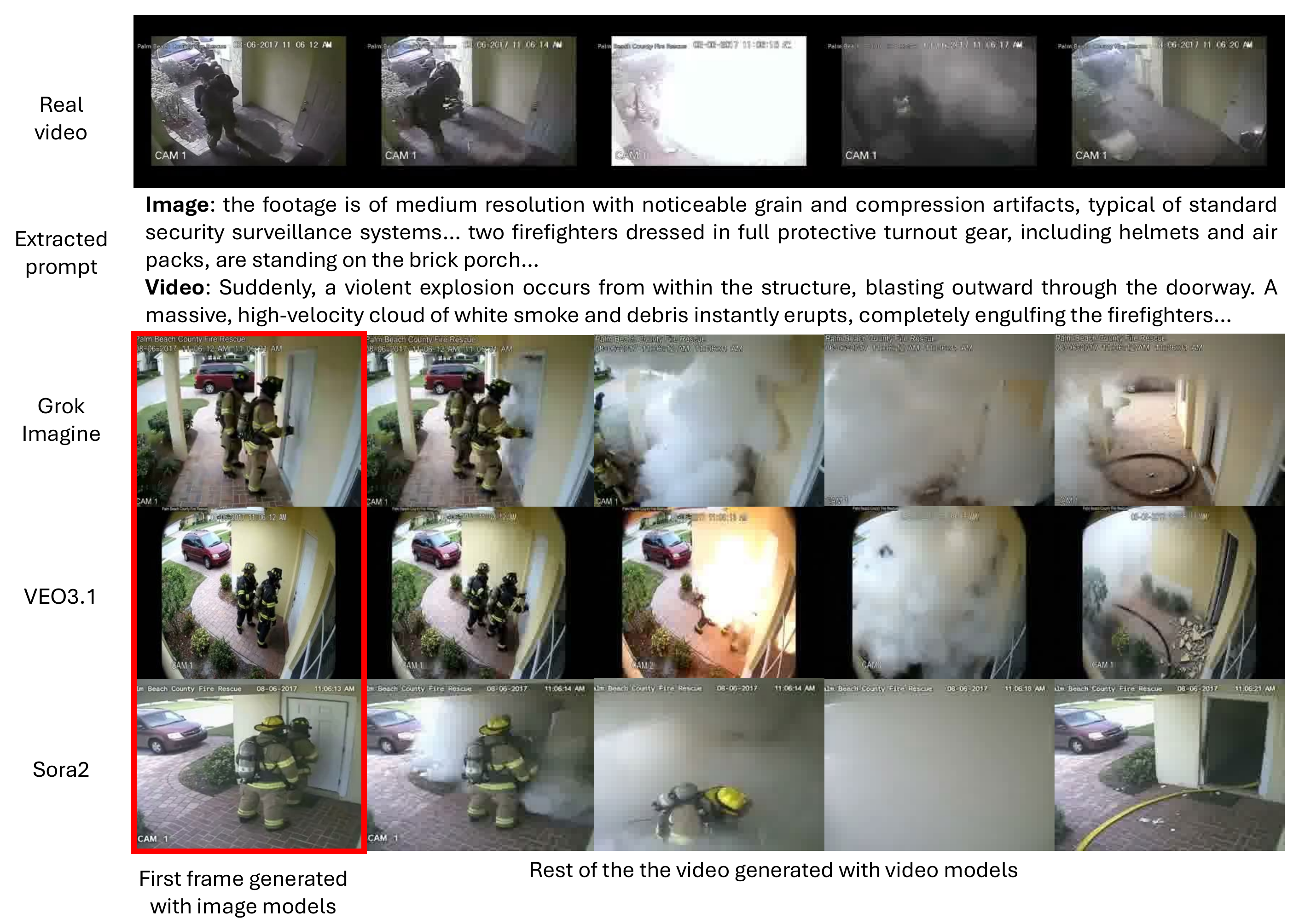}
   \caption{\textbf{Overview of dataset creation:} Based on a real video a prompt describing the content of the first frame as well as the entire video is extracted. Then an image model generates the first frame while a video model later generates the entire video based on the text prompt and the initial frame image. \textit{Note: the prompt shown in this figure is only an excerpt from the real prompt.}}
   \label{fig:overview}
\end{figure}

\section{Existing Datasets and Methods}

\subsection{Face-Based Deepfake Datasets}

Deepfake detection emerged from a person-centric setting where the perceived risk was the synthesis of identifiable individuals for impersonation, fraud, or non-consensual imagery.
Early datasets reflected this focus and concentrated on synthetic faces produced by face-swap and face-reenactment pipelines. The Deepfake Detection Challenge (DFDC) dataset~\cite{Dolhansky2020DFDC} popularized large-scale evaluation in this setting, releasing over 100{,}000 face videos generated by eight different methods.
Subsequent work expanded the diversity of generators tested.
The DF40 dataset~\cite{Yan2024df40} consolidates 40 open-source generators (face-swap, face-reenactment, and full face-synthesis pipelines) into a single evaluation suite designed to assess generalization across families of methods.
To better reflect deployment conditions, the WildDeepfake dataset~\cite{Zi2020wilddeepfake} collects deepfake videos directly from the web, capturing the compression artifacts, framing decisions, and subject diversity present in content that actually circulates online.
These datasets are important for face-centric deepfake detection but are not applicable to detection of fabricated \textit{events}.

\subsection{General-Content Video Deepfake Datasets}

As text-to-video and image-to-video models improved, detection datasets moved beyond faces toward arbitrary video content.
GenVideo~\cite{Chen2024genvideo} is a large dataset containing over two million videos using 16 open-source generators and a smaller set of 4 commercial systems (Pika, MoonValley, Sora, MorphStudio).
VidProM~\cite{wang2024vidprom} is an even larger dataset (6.7M videos)  from 3 open-source models and Pika Art.
The AIGVDBench dataset~\cite{Ma2026aigvd} expands the variety of generative models by combining 20 open-source and 11 closed-source models in a single benchmark.
Like CCDF, VidForensic~\cite{liu2025lavid} narrows the focus to commercial systems including videos from three generators (Sora, Kling, Gen-3) but at a much smaller scale (only 1{,}645 videos) and with the older commercial generators.
Similar to the CCDF setting, the SurFITR dataset~\cite{Wang2026surfitr} targets surveillance-style depictions of incidents and contains 137K samples produced by five open-source generators, but is restricted to still images and does not utilize the current generation of frontier commercial video generators.
Across these datasets, three limitations recur and complicate detector evaluation under realistic conditions: (1) most content is benign (animations, animal footage, fantastical scenes) rather than possibly malicious, (2) much of the synthetic material comes from older or open-source generators with visible artifacts (e.g., \cite{ma2026your} reports AUC=1.0 on the Open-Sora~\cite{Peng2025opensora2} generated videos), and (3) real and synthetic subsets can differ in resolution, frame rate, duration, and bitrate, allowing detectors to exploit non-content cues.
Table~\ref{tab:deepfake-datasets} summarizes how \name relates to other datasets.

\begin{table}[ht]
\centering
\caption{Comparison of select deepfake datasets. The ``T'' column describes the multimedia type, video (V) or image (I).}
\label{tab:deepfake-datasets}
\begin{tabular}{lclll}
\toprule
Dataset & T & \# Samples & Generators & Content Style \\
\midrule

DFDC~\cite{Dolhansky2020DFDC}          & V & 128K   & 8 Open-Source Models           & Face                \\
DF40~\cite{Yan2024df40}    & I/V & $>$1M   & 40 Open-Source Models          & Face                \\
WildDeepfake~\cite{Zi2020wilddeepfake}  & V & 707    & \textit{Unknown}                        & Face from Web       \\
\multirow{2}{*}{GenVideo~\cite{Chen2024genvideo}}      & \multirow{2}{*}{V} & \multirow{2}{*}{2.3M}  & 16 Open-Source Models, Pika,            & \multirow{2}{*}{General Web} \\
     &  &  & MoonValley, Sora, MorphStudio            &  \\
AIGVDBench~\cite{Ma2026aigvd} & V & 422K & 20 Open-, 11 Closed-Source Models & General Web \\
VidProM~\cite{wang2024vidprom}    & V & 6.7M  & 3 Open-Source Models, Pika art & General Web \\
VidForensic~\cite{liu2025lavid}& V & 1,645&Sora, Kling, Gen-3&  General Web\\
SurFITR~\cite{Wang2026surfitr}       & I & 137K   & 5 Open-Source Models             & Surveillance        \\
\midrule
\textbf{\name (Ours)}   & \textbf{V} & \textbf{1{,}840} & \textbf{Sora 2, VEO3.1, Grok Imagine}  & \textbf{Surveillance}        \\
\bottomrule
\end{tabular}
\end{table}



\subsection{Deepfake Detection}
\label{sec:methods}

Visual deepfake detection has evolved alongside the generators it targets.
Early methods were face-centric and exploited artifacts produced by face generation pipelines, including blending boundaries~\cite{Li2020xray}, frequency-domain inconsistencies~\cite{qian2020f3}, and facial identity information~\cite{Cozzolino2021idreveal}.
As generation capabilities developed to the point where arbitrary scenes could be produced more convincingly, detection broadened to more general content.
The ten detectors we benchmark on \name were all proposed in 2025 and represent the current state-of-the-art in this setting.
Table~\ref{tab:methods-overview} summarizes them.

The ten detectors fall into three broad approaches.
The first set of approaches use features from a pretrained model for classification.
C2P-CLIP~\cite{tan2025c2p} pairs each training image with a fixed ``real'' or ``fake'' text prompt and uses CLIP's image-text alignment to push the feature space toward better real-vs-fake separation.
ForgeLens~\cite{chen2025forgelens} adds lightweight modules into a frozen CLIP network to focus its features on forgery-relevant regions.
D3-Im~\cite{yang2025d} runs the image and a transformed copy (patches shuffled, flipped, or rotated) through CLIP, then a classifier compares both feature sets to predict real vs. fake.
The second set of approaches focus on the training data or augmentation.
B-Free~\cite{guillaro2025bias} and Community Forensics~\cite{park2025community} train a standard detector. B-Free on real images paired with their Stable Diffusion reconstructions to eliminate content and formatting biases.
Community Forensics on a large pool from 4{,}803 generators.
WaveRep~\cite{corvi2025waverep} replaces the low-frequency wavelet bands of fake training samples with those from paired real samples to push the detector to rely on mid-high frequency bands.
The third set of methods extracts dynamic video features.
L3DE~\cite{chang2025l3de} trains a 3D ConvNet on 3D cues (motion, depth, appearance) to flag failures of 3D physical consistency.
NSG-VD~\cite{zhang2025NSGVD} computes a physics-derived gradient via a pretrained diffusion model and then flags synthetic videos by measuring how different their gradient distribution is from a reference distribution of real videos.
D3-Vid~\cite{zheng2025d3vid} is training-free and uses second-order temporal differences of frame features to distinguish real and fake videos. STALL~\cite{ben2026training} is another training free approach that models spatial and temporal elements within a probabilistic framework.

\begin{table}[ht]
\centering
\caption{Overview of the nine detectors used to benchmark \name. The ``T'' column describes the input type: image (I) or video (V), while the ``Approach'' column describes the broad categories: pre-trained (PT), training data (TD), or dynamic video features (DV).\label{tab:detectors}}
\label{tab:methods-overview}
\begin{tabular}{lcccl}
\toprule
\textbf{Method} & \textbf{T} & \textbf{Approach} & \textbf{Year} & \textbf{Synthetic Training Data Type} \\
\midrule
C2P-CLIP~\cite{tan2025c2p}            & I & PT & 2025 & GAN + diffusion images\\
ForgeLens~\cite{chen2025forgelens}    & I & PT & 2025 & GAN images\\
D3-Im~\cite{yang2025d}                & I & PT & 2025 & GAN + diffusion images \\
B-Free~\cite{guillaro2025bias}        & I & TD & 2025 & Diffusion images \\
Community-Forensics~\cite{park2025community}  & I & TD & 2025 & Diffusion images \\
\midrule
WaveRep~\cite{corvi2025waverep}       & V & TD & 2025 & Pyramid flow videos \\
L3DE~\cite{chang2025l3de}             & V & DV & 2025 & Diffusion videos \\
NSG-VD~\cite{zhang2025NSGVD}          & V & DV & 2025 & Diffusion videos \\
D3-Vid~\cite{zheng2025d3vid}          & V & DV & 2025 & \textcolor{gray}{[NONE]} \\
STALL~\cite{ben2026training}          & V & DV & 2026 & \textcolor{gray}{[NONE]} \\
\bottomrule
\end{tabular}
\end{table}

\section{\name dataset}


Toward improving existing deepfake detection for non-face video content, we construct a new dataset, CCtv DeepFakes, referred to as \name.
\name fills two gaps in current deepfake detection datasets: (1) it uses frontier commercial generative models and (2) it depicts higher-stakes events (\textit{e.g.}, arson) rather than ``benign'' data (\textit{e.g.}, dog pictures). \name is designed as a small scale evaluation set containing highly realistic data, with each video being based on human verified description of actual CCTV footage, while limiting biases, related to resolution, duration or framerate, often present in commonly used datasets.
Figure \ref{fig:overview} shows an overview of dataset construction.
The choice of surveillance footage stems from two observations: (1) people are less likely to doubt generated surveillance footage as they expect this type of content to be of lower quality and/or contain artifacts and (2) such generated footage has already been widely shared while presented as real online\footnote{\tiny\url{https://phys.org/news/2025-11-ai-generated-wildlife-videos-generate.html}}. Unlike previous datasets like GenVideo, which contain categories such as  ``animals'' or  ``plants,'' \name contains potentially malicious content that would warrant using a deepfake detection tool to verify its validity.
This surveillance setting more closely mimics real world usage likely to be encountered by deepfake detectors.


\subsection{Prompt Creation}

To generate realistic surveillance footage, we base the prompts on 3 surveillance footage datasets widely used for anomaly detection.
Specifically, UCF-Crime \cite{sultani2018real}, XD-Violence \cite{Wu2020not} and MSAD \cite{msad2024}. The use of three different datasets allows for higher diversity in content and format. 
We keep all the categories from UCF-Crime and MSAD but only keep ``Road Accident'', ``Explosion'' and ``Fighting'' from XD-Violence as it contains many non-surveillance videos (\textit{e.g.},~from movies and news reports).
We randomly select 666 videos from this set and manually cut out 20s clips corresponding to the event category.
We then generate prompts by imputing the videos to the ``gemini-3-pro-preview'' model.
The model is prompted\footnote{The full prompt used and other parameters are available in the Appendix.} to extract the following information from the clips.
\begin{itemize}
    \item \uline{Video Source and Technical Details}: description of the footage type (CCTV, dash-cam, doorbell camera, \textit{etc}.), resolution, sharpness, metadata overlay (\textit{e.g.}, logo or timestamp), or any other information that can be used to describe the footage not directly related to the action of the video.
    \item \uline{Background and Setting}: description of the environment, location, as well as lighting and colors.
    \item \uline{Initial Frame Description}: precise description of the first frame content, including people and objects.
    \item \uline{Actions and Events}: description of the main action of the video as a sequence of events and the motion of the camera (if any).
    \item \uline{Audio Description}: description of  sounds, noises, music, or voices (if any).
\end{itemize}
The generated prompts are manually reviewed and corrected by human annotators to ensure that descriptions match the actual source video content.
A full 20s clip is used to create the video description instead of 8s clip as most of the events (\textit{e.g.}, robbery or burglary) tend to last longer than 8s. 
A truncated clip would not allow the capture of all the meaningful information from the video.

\subsection{Video Generation}

We generate videos following a two step process.
First, we generate an image using the ``Video Source and Technical Details'', ``Background and Setting'' and ``Initial Frame Description'' sections from the textual video description.
Second, we use the generated image along with the ``Actions and Events'' and ``Audio Description'' sections of the textual video description to generate a video.
We employ this two-stage approach after preliminary tests showed it resulted in higher quality videos than a direct text-to-video approach.

We select three commercial providers to generate videos, OpenAI, xAI, and Google, based on three criteria: (1) performance, (2) ubiquity, and (3) accessibility.
On performance, Veo 3.1 tops the VBench 2.0~\cite{Zheng2025vbench2} total-score ranking as of January 2026\footnote{\tiny\url{https://huggingface.co/spaces/Vchitect/VBench_Leaderboard}}.
Grok Imagine and Veo 3.1 occupy the top two spots on the human preference leaderboard in the Artificial Analysis Image-to-Video Arena (with audio) ranking as of January 2026\footnote{\tiny\url{https://artificialanalysis.ai/video/leaderboard/image-to-video?audio-output=true}}~\cite{Chiang24cbarena}.
On prevalence, while now closed, the Sora companion app surpassed 20.4 million downloads within five months of release~\cite{appmagic2026sora}, xAI claimed Grok Imagine generated 1.245 billion videos in the 30 days prior to release of version 1.0\footnote{\tiny\url{ https://www.socialmediatoday.com/news/x-formerly-twitter-updated-grok-video-generator-xai-imagine/811147/}}, likely due to native integration in the X social media service, and Veo 3.1 is offered inside Adobe Firefly\footnote{\tiny\url{https://www.adobe.com/products/firefly.html}}, placing it within easy reach of professional editing pipelines even for users unfamiliar with generative AI. By using these frontier generator which are integrated into widely used software and websites, we more accurately simulate the quality of deepfakes generated by non-expert users. In contrast to existing datasets that focus on open source approaches that may be too complex to operate for the general public.

In the two stage video generation, we employ OpenAI GPT Image 1.5\footnote{\tiny\url{https://developers.openai.com/api/docs/guides/image-generation}} (gpt-image-1.5-2025-12-16), xAI Grok Imagine Image\footnote{\tiny\url{https://docs.x.ai/developers/model-capabilities/images/generation}} (grok-imagine-image-2026-03-02), and Google Nano Banana Pro\footnote{\tiny\url{https://deepmind.google/models/gemini-image/}}  (gemini-3-pro-image-preview) for image generation.
While for video generation, we use OpenAI Sora 2\footnote{\tiny\url{https://developers.openai.com/api/docs/guides/video-generation}} (sora-2-2025-12-08, with sora-2-pro-2025-10-06 used as a fallback for videos blocked at the sora-2-2025-12-08 moderation stage), Grok Imagine-Video\footnote{\tiny\url{https://docs.x.ai/developers/models/grok-imagine-video}}  (model version not provided, generation took place between  2026-04-08 and 2026-04-16) and Google VEO 3.1\footnote{\tiny\url{https://aistudio.google.com/models/veo-3}} (veo-3.1-generate-preview).
In this process, image and video model from the same provider are paired (\textit{e.g.}, Sora 2 video generation uses only images GPT Image 1.5). For VEO3.1, images were generated at resolution 1200$\times$896 and 1376$\times$768 with a 70\%/30\% split to mimic the original data aspect ratio distribution. Video were then generated at 1280$\times$720, VEO3.1 do not allow lower resolutions and only portrait or landscape aspect ratio. Grok images were generated at resolution 1200$\times$896 and 1408$\times$768 with a 70\%/30\% split to mimic the original data aspect ratio distribution. Videos were generated at resolution 640$\times$480 (images at 1200$\times$896) and resolution 1280$\times$720 (images at 1408$\times$768) to mimic the original data distribution. For Sora 2, images are generated at resolution 1024$\times$1024. Videos are then generated at resolution 1280$\times$720 with images scaled and padded to fit video dimensions (but retain aspect ratio). All generated videos for all models are 8s long, matching the maximum length that VEO3.1 can generate. All video are generated with audio enabled and feed the ``Audio Description'' section even if it simply described a lack of audio. More details about the settings can be found in the Appendix.

\subsection{Dataset Curation}
The video generation process results in 657 videos for Grok, 607 videos for Veo, and 501 videos for Sora 2.
These counts, lower than the original 666 videos, correspond to blocked requests by the moderation stage of the generative models when generating either images or videos.
To balance our dataset, we keep only videos that have successfully generated by all 3 models leaving us with 1{,}840 videos overall (460 real videos and 460 videos for each of the 3 generators).
To avoid trivial distinctions between real and generated videos, we provide a ``cleaned'' version of \name, designed to limit the differences between real data and generated data.
First we limit the duration of real video to 8s (some videos that are originally shorter are kept at their original duration), we set the frame rate to 24 fps for all videos, remove the audio track, and ensure the video codec is the same for all videos (H.264).
The cut to 8s was performed manually for all videos to keep the focus on the event category of the video.
As the source videos contain 16 different resolutions compared to the 2 available for generated videos, we mirror the resolution of generated videos onto their real counterparts.
Finally, as we observed a large difference in average bitrate between real and generated videos, we also set an upper bound on the bitrate of generated videos.
To simulate a low effort attack on detectors, we provide an ``altered'' version of \name.
The alterations consist of reducing the fps to 6, randomly modifying the saturation and hue value to alter the colors, and halving the resolution for 720p videos.
The modifications are performed with FFmpeg and take less than a second to modify a single video.

\subsection{Dataset Statistics}

Table \ref{tab:Dataset_stats_2} shows the selection from the base set of 666 videos.
We observe that event categories corresponding to violent behaviors (\textit{e.g.}, Robbery or Shooting) were more heavily moderated and this resulted in discarding up to two-thirds of the original videos in that event category.
We also observe that Sora 2 has the strictest moderation and blocked more than 150 videos.
But we also note that videos that pass the Sora 2 moderation could be moderated by the other two models.
In Table~\ref{tab:Dataset_stats_4} we summarize the video parameters of \name for the raw data (\textit{i.e.}, before cleaning) and our cleaned set.
We see that if we use the raw data when evaluating deepfake detection methods we introduce several important biases with regards to resolution, frame rate and bitrate.
This can lead to models relying on these superficial cues rather than more fundamental deepfake cues to make predictions. 
Table~\ref{tab:deepfake-datasets} compares \name to other widely used generated video datasets.
We note that, while \name contains less data than some of the other datasets, we focus only on recent generative methods that can produce convincing generated content and on the specific setting of surveillance footage, which is more likely to be used for harmful purposes.

%

\subsection{Distribution, Reproducibility, licensing and ethics}
CCDF is made available for research purpose by filling the form at \url{https://docs.google.com/forms/d/e/1FAIpQLScSmOnvIXIsAItzLosiUUpAWVnW7LeukjgODTeF4g3kv9eqJA/viewform}. The shared data includes only the raw generated videos from CCDF and the prompts used to generate them. To comply with the license agreement of UCF-Crime, XD-Violence and MSAD we do not redistribute the real data used in the experiments. We however provide code and instructions to reproduce the real data from the original datasets. We also provide file hashes to ensure the reconstructed data exactly matches the video that was used in our experiments. Furthermore, we provide code to recreate the Cleaned and Altered sets from our experiments. 
This dataset contains generated surveillance video footage generated using artificial intelligence. The content is provided solely for research, development, and educational purposes, such as computer vision training, anomaly detection research, and model evaluation. It must not be used in any context that could mislead others into believing the footage is real.

\begin{table}[ht]
    \centering
    
    \caption{\name breakdown \textit{w.r.t.} dataset, event category, and generative method.}
    \resizebox{0.65\linewidth}{!}{
    \begin{tabular}{ccccc|c}
        \toprule
        \textbf{Category} & \textbf{Real} & \textbf{Grok} & \textbf{VEO3.1} & \textbf{Sora2}& \textbf{Selection} \\
        \midrule
        \midrule
        UCF-Crime & 444 & 436 & 413 & 329 & 307\\
        \midrule
        Abuse & 24 & 20 & 21 & 20 &16 \\
        Arrest & 20 & 20 & 18 & 10 & 9 \\
        Arson & 29 & 29 & 29 & 19  & 19\\
        Assault & 39 & 37 & 36 & 25$^*$ &21\\
        Burglary & 61 & 61 & 61 & 50 &50 \\
        Explosion & 33 & 33 & 29 & 31 &27\\
        Fighting & 27 & 26 & 24 & 18$^*$ &16\\
        Road Accident & 20 & 19 & 18 & 20 &17 \\
        Robbery & 44 & 44 & 32 & 16 &13\\
        Shooting & 29 & 29 & 27 & 13 &12\\
        Shoplifting & 31 & 31 & 31 & 30 &30\\
        Stealing & 49 & 49 & 49 & 40 &40\\
        Vandalism & 38 & 38 & 38 & 37&37 \\
        \midrule
        \midrule
        XD-Violence & 54 & 54 & 43 & 42&36 \\
        \midrule
        Explosion & 31 & 31 & 20 & 21 &15 \\
        Fighting & 2 & 2 & 2 & 0 &0\\
        Road Accident & 21 & 21 & 21 & 21&21 \\
        \midrule
        \midrule
        MSAD  & 168 & 167 & 151 & 130& 117 \\
        \midrule
        Assault & 10 & 10 & 9 & 3 &3 \\
        Explosion & 10 & 10 & 8 & 8&6 \\
        Fighting & 8 & 8 & 8 & 2 &2\\
        Fire  & 19 & 19 & 19 & 19$^*$&19 \\
        Object Falling  & 16 & 16 & 16 & 16&16 \\
        People Falling  & 30 & 30 & 28 & 28 &26\\
        Road Accident & 26 & 26 & 25 & 26&25 \\
        Robbery & 21 & 21 & 16 & 10 &8\\
        Shooting & 12 & 11 & 9 & 6 &3\\
        Vandalism & 10 & 10 & 9 & 6 &5\\
        Water Incident & 6 & 6 & 4 & 6 &4\\
        \midrule
        \midrule
        Total & 666 & 657 & 607 & 501&460 \\
        \bottomrule
        \multicolumn{6}{c}{\footnotesize `$^*$' denotes 1 video in group generated with sora-2-pro-2025-10-06 and selected for cleaned set} 
    \end{tabular}}
    \label{tab:Dataset_stats_2}
    
\end{table}

\begin{table}[ht]
    \centering
    \caption{\name metadata summary for raw data and the cleaned set.}
    \resizebox{0.85\linewidth}{!}{
    \begin{tabular}{cccccc}
        \hline
        \textbf{Data} & \textbf{number} & \textbf{avg duration} & \textbf{fps} & \textbf{resolution}& \textbf{avg bitrate} \\
                \hline
                 \multicolumn{6}{c}{\textbf{Raw data}} \\
                \hline
        \textbf{Real data} & 460 & 7.94 & 6 to 30 & 320$\times$240 to 1920$\times$1080  & 389.58 \\
        \textbf{Grok} & 460 & 8.04 & 24 & 640$\times$480 and 1280$\times$720  &  3036.63 \\
        \textbf{VEO} & 460 & 8.00 & 24 & 1280$\times$720  &  2335.15 \\
        \textbf{Sora2} & 460 & 8.30 & 30 & 1280$\times$720  & 4295.10 \\
                \hline
                 \multicolumn{6}{c}{\textbf{Cleaned data}} \\
                \hline
        \textbf{Real data} & 460 & 7.94 & 24 & 320$\times$240 to 1920$\times$1080  & 355.21 \\
        \textbf{Grok} & 460 & 8.00 & 24 & 320$\times$240 to 1920$\times$1080  &  366.95 \\
        \textbf{VEO} & 460 & 8.00 & 24 & 320$\times$240 to 1920$\times$1080  &  367.15 \\
        \textbf{Sora2} 0& 460 & 8.00 & 24 & 320$\times$240 to 1920$\times$1080  & 372.21 \\
        \hline
    \end{tabular}}
    \label{tab:Dataset_stats_4}
\end{table}

\section{Benchmarks}

We evaluate \name on 10 methods described in Section~\ref{sec:methods}.
We evaluate these methods on the cleaned and altered versions of the \name dataset to assess how simple alterations can affect detection models.
We use the following 7 metrics for evaluation, following the most commonly used metrics for deepfake detectors: Area Under the Curve (AUC), Average Precision (AP), Balanced Accuracy (Acc) with a threshold at 0.5, recall, F1-Score, Expected-Calibration-Error (ECE), Negative Log Likelihood (NLL).
For each method we use the code and weights provided by the authors on their github pages.
None of the methods are retrained on data of \name and tested in a zero-shot setting, simulating an operational setting where the detector must classify real or fake given only a video.
For image-based methods, we extract frames at 1 fps (8 images per video) and compute the score on a per video basis by averaging the logits returned by the network for each set of 8 images.

Table \ref{tab:benchmark} presents the AUC, AP and Acc values on the cleaned setting while \ref{tab:benchmark_raw} present the results on the raw data from the generators for comparison.
Full results for the raw, cleaned and altered sets are presented in the Appendix in tables \ref{tab:raw}, \ref{tab:image_cleaned} and \ref{tab:image_altered}, respectively.
We choose to focus on those three metrics since every methods tested reports at least one of those. In the Appendix we provide an ablation study of the effect of each of the modification applied to obtain the cleaned set.
Additionally, we report in Table \ref{tab:genvideo} (full results in Appendix Tables \ref{tab:genvideo_AUC} to \ref{tab:genvideo_NLL}) the performance of every method on the test set from GenVideo \cite{Chen2024genvideo}, the most widely used dataset in the literature on video deepfake detection. This additional benchmark allow us to ensure that \textbf{1)} The code used in the experiment is representative of the ability of each method. \textbf{2)} That images based approaches work properly on frames extracted from video feeds. \textbf{3)} that the 0.5 threshold for accuracy is correct (only a few methods could benefit from another value). \textbf{4)} We have a proper frame of reference to analyze results on CCDF.

On the Cleaned set, we observe that none of the 10 methods accurately distinguish between real and generated \name videos.
Specifically many methods either classify nearly all videos as real or fake as indicated by the recall values, leading to discrepancies between metrics.
This is in stark contrast with the results on GenVideo.
This can be explained by several factors, the likely primary reason, shared by all methods, is the training data used.
The training data is  dated for many of the methods (\textit{i.e.}, C2P-CLIP, Forgelens, D3-Im, C-forensic and  D3-vid  only use generators from 2023 or before).
For the others that use more recent training data (2024), the generators are open-source and significantly less effective than frontier commercial models and generate obvious deepfake artifacts. 
Furthermore, many methods also use real data from older sources (\textit{e.g.}, ImageNet~\cite{5206848}) leading to a focus on low resolution real data while the generated data will typically be higher resolution.
While \name contains resolution as low as 320$\times$240, which is still higher than some of these datasets, we also provide higher resolution content making the dataset more diverse and potentially more challenging for these models.
Finally, most of the data used for training presents large differences between real and generated videos in term of resolution, frame rate or duration.
For example, in GenVideo~\cite{Chen2024genvideo} training set, real videos have a resolution between 224$\times$224 and 640$\times$340 with duration between 5s and 120s.
Meanwhile, the generated  videos have a resolution between 512$\times$512 and 1280$\times$720 and duration between 2s and 6s.
Contrary to this, \name provides a cleaned set where the video properties are the same for real and generated videos, making it harder for model who might have learned these biases.
On top of these, B-Free is trained on a dataset of edited images instead of fully generated which can also increase the evaluation gap.

The results on the raw set shows that only Community-Forensics and Waverep are able to obtain moderate detection performance on the raw data of \name. However, the detection performance still drop significantly compared to Genvideo. This indicates that both methods have better out of distribution capabilities than the 8 other approaches. However their performances drop considerably when evaluated on the Cleaned set, indicating that they rely at least partly on data distribution biases to make decision, making them vulnerable to simple attacks. While the drop can be of varying intensity it is observed for all methods, this highlight the need to limit biases when developing new dataset. 

Overall, the benchmark shows that recent methods struggle in a zero-shot setting on CCDF. We attribute this for the most part to the use of frontier generative models in CCDF. However, the setting of CCTV footage is also important as it allows us to simulate deepfakes that would require the use of detectors. \cite{ben2026training} report AUC and AP of 0.86 and 0.87 respectively on VEO3 generated data, however in our experiments on CCDF show that even using the raw data, STALL only achieve 0.599 AUC and 0.587 AP on VEO3.1. This highlight the fact that beyond the use of frontier models, the surveillance settings footage also prove more challenging.

The detection results on the altered version likewise performs poorly (see  Table~\ref{tab:image_altered} in the Appendix).
The purpose of the altered version is to evaluate if detectors that performed well on the cleaned version would be resilient to simple alterations to the videos.
However, since no detector performed well on the cleaned set, we cannot evaluate this hypothesis.
We will distribute the altered version of the \name dataset alongside the cleaned version as it could be useful for future research with better detectors.

This study highlights two important points. First, the need for  deepfake detection datasets that match high-stakes operational scenarios.
Second, the need for more robust detection methods that can accurately detect generated content from frontier video generators.

    \begin{table}
        \centering
        \small
        \caption{Evaluation of detectors on \name cleaned set for Area Under the Curve; Average Precision and Accuracy.}
        \resizebox{1.0\linewidth}{!}{
        \begin{tabular}{lcccc}
            \hline
            \textbf{Method} & Grok & VEO3.1 & Sora2 & Global \\
            \hline
                 \multicolumn{5}{c}{\textbf{AUC / AP / Acc}} \\
            \hline
            B-Free\cite{guillaro2025bias} & 0.427 / 0.473 / 0.460 & 0.359 / 0.401 / 0.359& 0.465 / 0.467 / 0.465& 0.417 / 0.704 / 0.450\\
            C2P-CLIP\cite{tan2025c2p} & 0.409 / 0.426 / 0.510&  0.419 / 0.430 / 0.502& 0.402 / 0.424 /0.504& 0.410 / 0.684 / 0.505\\
            ForgeLens\cite{chen2025forgelens} & 0.322 / 0.411 / 0.500& 0.327 / 0.400 / 0.500& 0.399 / 0.440 / 0.500 & 0.349 / 0.416 / 0.500\\ 
            D3-Im\cite{yang2025d}& 0.510 / 0.542 / 0.508 & 0.621 / 0.630 / 0.511 & 0.469 / 0.521 / 0.510& 0.533 / 0.790 / 0.509\\
            C-Forensics\cite{park2025community}& 0.711 / 0.710 / 0.505& 0.693 / 0.677 / 0.502& 0.750 / 0.735 / 0.507 & 0.718 / 0.875 / 0.504\\
            \hline
            D3-Vid\cite{zheng2025d3vid}& 0.506 / 0.480 / 0.540& 0.523 / 0.490 / 0.547& 0.543 / 0.532 / 0.504& 0.524 / 0.749 / 0.530\\
            L3DE\cite{chang2025l3de} & 0.663 / 0.630 / 0.588 & 0.701 / 0.661 / 0.692 & 0.602 / 0.601 / 0.575& 0.665 / 0.835 / 0.588\\
            NSG-VD\cite{zhang2025NSGVD} & 0.490 / 0.503 / 0.486& 0.550 / 0.553 / 0.543& 0.463 / 0.457 / 0.471& 0.501 / 0.752 / 0.500\\
            WaveRep\cite{corvi2025waverep} & 0.684 / 0.653 / 0.504& 0.792 / 0.804 / 0.558 & 0.592 / 0.589 / 0.506& 0.689 / 0.865 / 0.522\\
            STALL\cite{ben2026training} &0.592 / 0.574 / 0.546 & 0.547 / 0.538 / 0.525 & 0.635 / 0.618 / 0.584 & 0.591 / 0.577 / 0.551 \\
            \hline
            
        \end{tabular}}
            \label{tab:benchmark}
    \end{table}

    \begin{table}
        \begin{minipage}[t]{0.45\linewidth}

        \centering
        \small
        \caption{Evaluation of detectors on \name raw set for Area Under the Curve; Average Precision and Accuracy.}

        \begin{tabular}{lc}
            \hline
            \textbf{Method}  & Global \\
            \hline
                 \multicolumn{2}{c}{\textbf{AUC / AP / Acc}} \\
            \hline
            B-Free\cite{guillaro2025bias} & 0.578 / 0.790 / 0.563\\
            C2P-CLIP\cite{tan2025c2p}& 0.548 / 0.778 / 0.510 \\
            ForgeLens\cite{chen2025forgelens} & 0.702 / 0.723 / 0.506\\ 
            D3-Im\cite{yang2025d}& 0.670/ 0.866 / 0.526\\
            C-Forensics\cite{park2025community} & 0.789 / 0.913 / 0.515\\
            \hline
            D3-Vid\cite{zheng2025d3vid}& 0.590 / 0.802 / 0.534\\
            L3DE\cite{chang2025l3de} & 0.256 / 0.621 / 0.384\\
            NSG-VD\cite{zhang2025NSGVD} & 0.689 / 0.844 / 0.605\\
            WaveRep\cite{corvi2025waverep} & 0.882 / 0.961 / 0.720\\
            STALL\cite{ben2026training}  & 0.631 / 0.618 / 0.576  \\
            \hline
            
        \end{tabular}
            \label{tab:benchmark_raw}
                \end{minipage}%
                \hfill%
    \begin{minipage}[t]{0.45\linewidth}
        \centering
        \small
        \caption{Evaluation of detectors on GenVideo test set for Area Under the Curve; Average Precision and Accuracy.}
        
        \begin{tabular}{lc}
            \hline
            \textbf{Method} & Global \\
            \hline
                 \multicolumn{2}{c}{\textbf{AUC / AP / Acc}} \\
            \hline
            B-Free\cite{guillaro2025bias} & 0.987 / 0.990 / 0.948\\
            C2P-CLIP\cite{tan2025c2p}& 0.897 / 0.903 / 0.514\\
            ForgeLens\cite{chen2025forgelens} & 0.701 / 0.530 / 0.667\\ 
            D3-Im\cite{yang2025d}& 0.954 / 0.960 / 0.880\\
            C-Forensics\cite{park2025community} & 0.973 / 0.981 / 0.952\\
            \hline
            D3-Vid\cite{zheng2025d3vid}& 0.972 / 0.962 / 0.923\\
            L3DE\cite{chang2025l3de} & 0.502 / 0.463 / 0.486\\
            NSG-VD\cite{zhang2025NSGVD} & 0.938 / 0.931 / 0.866\\
            WaveRep\cite{corvi2025waverep} &0.982 / 0.987 / 0.933\\
            STALL\cite{ben2026training} & 0.704 / 0.693 / 0.522\\
            \hline

        \end{tabular}
            \label{tab:genvideo}
            \end{minipage}
    \end{table}

\section{Limitations}

Despite its advantage over existing datasets, \name suffers from some limitations that will be addressed in future work.
First, while we curated the content of the dataset, we did not evaluate if each generated video reliably fools people into mistaking them for real content.
Future works will include an user study to rank videos based on their ability to deceive people. 
Second, the size of \name is still relatively small compared to web-scraped datasets and is limited to 3 methods.
Future work will increase the size of the dataset while including additional generative video models. However, \name will remain an evaluation-only benchmark and is not intended to reach the scale required for training detection algorithms.
Specifically, we will modify video descriptions to be more moderation compliant without altering their content, thereby reducing the number of videos that are not generated. 
Third, due to generative models limitations, video length is still short (8s) while real surveillance footage is likely longer.
Lastly, \name currently only includes 16 action categories, some of them with less than 10 valid samples which is not entirely representative of the diversity of content captured by surveillance camera.
Future versions of \name will include more varied video length, particularly longer video if the evolution of generative models permit it and extend the number of action categories.

\section{Summary}

We propose \name a new dataset of generated videos that better represents  real-world deepfakes by focusing on surveillance footage which is more likely to be used for malicious purposes and only uses recent commercial generators.
The results of our benchmark demonstrate the inability of current detection methods to distinguish generated video from real video, despite their good performance on other datasets.
This shows that current state-of-the-art open-source detectors can not be used in real-world scenarios and the difficulty in protecting users from generated content using automated approaches.
A likely reason behind the failure of these detector models is the data used to train and evaluate them, as they all perform well on other datasets which focus on benign content and older generative methods.
This highlights two important challenges: (1) the need for more difficult generated video datasets representing more malicious attacks and (2) the need for detection methods that are more resilient to zero-shot detection on recent methods.
With \name, we make one step towards solving the first challenge and we hope that its release will help the deepfake detection community solve the second challenge.
Future works will focus on improving \name to better represent the quality and diversity of malicious deepfakes used in real scenarios.

\section{Acknowledgment}
This research work has been funded by the German Federal Ministry of Education and Research and the Hessian Ministry of Higher Education, Research, Science and the Arts within their joint support of the National Research Center for Applied Cybersecurity ATHENE.

\FloatBarrier

\small

\bibliographystyle{abbrv}
\bibliography{Bibliography}

\newpage

\appendix

\section{Technical appendices and supplementary material}

\subsection{Disclaimer}

This dataset contains generated surveillance video footage generated using artificial intelligence. The content is provided solely for research, development, and educational purposes, such as computer vision training, anomaly detection research, and model evaluation. It must not be used in any context that could mislead others into believing the footage is real.

\subsection{Prompt Generation Parameters}
In this section we describe the exact parameters used to extract prompts from real videos
using "gemini-3-pro-preview". Below is the exact prompt used, separated in two parts due to length. Additionally, the following system instruction was added each time to ensure strict rule following and avoid moderation: "\textit{You are a professional security and safety analysis AI. Your purpose is to provide objective, clinical, and forensic descriptions of surveillance footage. You are authorized to describe incidents of violence, accidents, and emergencies for documentation purposes. Do not self-censor objective facts. Avoid emotional language and focus on the mechanics of the events.}". The parameters "\textit{HARM\_CATEGORY\_DANGEROUS\_CONTENT}", "\textit{HARM\_CATEGORY\_HATE\_SPEECH}", "\textit{HARM\_CATEGORY\_HARASSMENT}" where set to \textbf{OFF} to further prevent moderation from blocking request. All other parameters are kept to their default values. We feed each video accompanied by the prompt to independent API instance to avoid memory leak from one video to the other.

\begin{tcolorbox}[title={gemini-3-pro-preview Prompt Part 1}, colback=gray!3, colframe=black]

You are tasked with describing a video in a strictly factual and detailed manner. 
Do not make assumptions, guesses, or interpretations beyond what is directly visible or audible. 
The videos you will be shown always belong to one of the following categories: 
Vandalism, stealing, shoplifting, shooting, robbery, road accidents, fighting, explosion, burglary, assault, arson, arrest, abuse (implied of vulnerable people), object falling, water incident, fire (not arson), or people falling. 
Use this knowledge only to guide the precision of your description.

Your description must be based solely on the actual content of the video. 
If metadata is present (such as timestamps, camera identifiers, overlays, or labels), note it explicitly in the description.

Interpretation is allowed only when it is necessary to describe the **main action of the video** that falls within the categories above. 
In such cases, you may state the action (e.g., "the person is shoplifting" or "a fight occurs"), but you must not speculate about motives, identities, or unseen elements. 
All other aspects of the description must remain strictly factual and observational.

\subsection*{Style Requirements}
\begin{itemize}[leftmargin=1.2em]
    \item Stay factual and concise.
    \item Interpret only the main action when necessary, and only within the given categories.
    \item Do not speculate about identities, intentions, or unseen elements.
    \item Use neutral, objective language.
    \item Explicitly avoid subjective judgments or emotional framing.
\end{itemize}

Your description must include the following sections with the same formatting for the 4 parts, do not add any comment outside of these in your answer, also do not finish with a question:

\end{tcolorbox}

\begin{tcolorbox}[title={gemini-3-pro-preview Prompt Part 2}, colback=gray!3, colframe=black]

\subsection*{1. \textbf{Video Source and Technical Details}}
Presented with three keypoints:
\begin{itemize}[leftmargin=1.2em]
    \item Describe the video quality (resolution, clarity, sharpness)
    \item Mention the camera position, angle, zoom, and stability.
    \item Note if the video appears handheld, fixed, or from a specific perspective.
    \item Include any metadata overlays (timestamps, camera labels, identifiers).
    \item These part should be presented with three keypoints (video Quality, camera Details, metadata)
\end{itemize}

\subsection*{2. \textbf{Background and Setting}}
\begin{itemize}[leftmargin=1.2em]
    \item Describe the environment, location, objects, and context visible in the background.
    \item Note lighting, colors, and any static elements. 
    \item This part should be presented as a single paragraph.
\end{itemize}

\subsection*{3. \textbf{Initial Frame Description}}
\begin{itemize}[leftmargin=1.2em]
    \item Describe the very first frame of the video in detail. 
    \item Include visible objects, people, lighting, and any metadata overlays.
    \item This initial frame can be used to help with understanding the overall context and progression of the video.
    \item This part should be presented as a single paragraph.
\end{itemize}

\subsection*{4. \textbf{Actions and Events}}
Presented with two keypoints:
\begin{itemize}[leftmargin=1.2em]
    \item Describe all movements, gestures, and actions taking place in detail.
    \item Specify the sequence of events clearly and factually.
    \item Do not give timestamps in this section.
    \item Avoid describing blood, open wounds, or specific physical injuries.
    \item If necessary, interpret the main action in terms of the categories above, but remain neutral and objective.
    \item It is not obligatory to explicitely state the category name (e.g The video depicts an incident of abuse), do it only when it helps the description and naturally in the description when the event happens.
    \item Describe again camera motion if any. Emphasis the lack of camera motion when there is none. 
    \item This part should be presented with two keypoints: Action described in a single paragraph and camera motion also described in a single paragraph.
\end{itemize}

\subsection*{5. \textbf{Audio Description}}
\begin{itemize}[leftmargin=1.2em]
    \item Describe voices, sounds, music, or ambient noise. 
    \item Include tone, volume, clarity, and any background sounds.
    \item This part should be presented as a single paragraph.
\end{itemize}

\end{tcolorbox}

\begin{tcolorbox}[title={Example of extracted prompt (after human curation)}, colback=gray!3, colframe=black]

1. Video Source and Technical Details

- **Video Quality:** The video resolution is low, resulting in a grainy and pixelated image typical of older surveillance systems.

   - **Camera Details:** The footage is captured from a fixed, high-angle security camera positioned to overlook a street and sidewalk. The view is static with no zoom or panning.
   
   - **Metadata:** A timestamp overlay is visible in the top-left corner, but it is not readable.

2. Background and Setting
The scene is set on a paved residential street during the daytime. On the left side of the frame, a concrete sidewalk is bordered by brown, dormant grass, suggesting a late autumn or winter season. A black sedan is parked along the curb on the left, facing towards the camera. To the right of the road, there is a patch of grass and dirt. The lighting is diffuse, indicating overcast weather conditions.

3. Initial Frame Description
The first frame shows the black sedan parked on the left side of the street with its front amber lights illuminated. The vehicle is stationary and faces the camera. The road is otherwise empty, and the surrounding area appears quiet.

4. Actions and Events

- **Action described:** A dark-colored minivan enters the frame from the top, driving down the road towards the camera. It slows down and pulls up directly alongside the parked black sedan. The sliding door opens up a little and someone holds a gun out of the the open door, shooting into the the parked sedan. The minivan stays in motion. The event can be described as a drive by.

   - **Camera motion:** The camera remains completely stationary throughout the entire incident, with no panning, tilting, or zooming adjustments.

5. Audio Description
There is no audio.

\end{tcolorbox}

\subsection{Gemini and VEO3.1 Parameters}
\textbf{Image generation}: We generate images using "gemini-3-pro-image-preview" with \textit{candidate\_count=1} and \textit{aspect\_ratio=16:9 or 4:3} as explained in the main paper. Additionally, the parameters "\textit{HARM\_CATEGORY\_DANGEROUS\_CONTENT}", "\textit{HARM\_CATEGORY\_HATE\_SPEECH}", "\textit{HARM\_CATEGORY\_HARASSMENT}" where set to \textbf{OFF} to further prevent moderation from blocking request. All other parameters are kept to their default values.

\textbf{Video generation}: We generate videos using "veo-3.1-generate-preview" with parameters \textit{number\_of\_videos=1}, \textit{aspect\_ratio=16:9} ,\textit{resolution=720p} and every other parameters kept at their default values.

For both image and video generation we use Gemini API and use a separate instance for each generation to avoid memory leak.

\subsection{Grok Imagine Parameters}
\textbf{Image generation}: We generate images using "grok-imagine-image" with \textit{resolution=1k} and \textit{aspect\_ratio=16:9 or 4:3} as explained in the main paper.

\textbf{Video generation}: We generate videos using "grok-imagine-video" with parameters \textit{duration=8}, \textit{aspect\_ratio=16:9 or 4:3} and \textit{resolution=480p or 720p} as explained in the main paper and every other parameters kept at their default values.

For both image and video generation we use xAI API and use a separate instance for each generation to avoid memory leak.

\subsection{Sora 2 Parameters}
\textbf{Image generation}: We generate images using "gpt-image-1.5-2025-12-16" with \textit{size=1024$\times$1024}, \textit{quality=high}, \textit{moderation=low} as explained in the main paper.

\textbf{Video generation}: We generate videos using "sora-2-2025-12-08" with "sora-2-pro-2025-10-06" used as a fallback for videos blocked at the sora-2-2025-12-08 moderation stage. For either model, parameters are set to \textit{seconds=8}, \textit{size=1280$\times$720}.

For both image and video generation, we use the OpenAI API and separate instances for each generation prompt to prevent interference among the prompts.

\subsection{Additional Statistics}
In Table~\ref{tab:Dataset_stats_3} we provide a category based breakdown of \name without regards for the source dataset. We observe that there is a imbalance in the original data but also that some categories are smaller due to the moderation of generative which exacerbates the inter-category imbalance. Future versions of \name will improve this aspect.

\begin{table}[ht]
    \centering
    \caption{Statistics of the generated data}
    \begin{tabular}{ccccc|c}
        \hline
        \textbf{Base dataset} & \textbf{Real} & \textbf{Grok} & \textbf{VEO3.1} & \textbf{Sora2}& \textbf{Selection} \\
        \hline

        Abuse & 24 & 20 & 21 & 20 &16\\
        Arrest & 20 & 20 & 18 & 10 &9\\
        Arson / Fire & 48 & 48 & 48 & 38 &38\\
        Assault & 49 & 47 & 45 & 28 &24\\
        Burglary & 61 & 61 & 61 & 50 &50\\
        Explosion & 74 & 74 & 57 & 60 &48\\
        Fighting & 37 & 36 & 34 & 20 &18\\
        Road Accident & 67 & 66 & 64 & 67&63 \\
        Robbery & 65 & 65 & 48 & 26 &21\\
        Shooting & 41 & 40 & 36 & 19 &15\\
        Shoplifting & 31 & 31 & 31 & 30 &30\\
        Stealing & 49 & 49 & 49 & 40 &40\\
        Vandalism & 48 & 48 & 47 & 43 &42\\
        Object Falling  & 16 & 16 & 16 & 16&16 \\
        People Falling  & 30 & 30 & 28 & 28 &26\\
        Water Incident & 6 & 6 & 4 & 6 &4\\
        \hline
        \hline
        Total & 666 & 657 & 607 & 501&460 \\

    \end{tabular}
    \label{tab:Dataset_stats_3}
\end{table}

\subsection{Full results}

In this section we present the full results for the table shown in the main paper

In table \ref{tab:raw} and \ref{tab:image_cleaned} we present the results of the nine metrics used for the evaluations of detection models on the raw and cleaned sets respectively. These results further show that no methods is able to accurately detect the deepfakes of \name. 

The full results on the altered set are presented in table \ref{tab:image_altered}. We do not present the results of STALL as the method do not work on videos with a framerate lower than 8fps while the altered set only contain video at 6fps. We observe little difference with the results on the cleaned set due to the poor performance of the detectors on this latter. 

In tables \ref{tab:genvideo_AUC}, \ref{tab:genvideo_AP}, \ref{tab:genvideo_ACC}, \ref{tab:genvideo_RECALL}, \ref{tab:genvideo_F1}, \ref{tab:genvideo_ECE}, \ref{tab:genvideo_NLL} we present the full results on GenVideo. This allow a frame of reference to compare against the results on CCDF.


\subsection{Ablation on the modification of the Cleaned set}

In this section we present the effect of the different modification made to create the Cleaned set from the raw set. For each experiment only one feature was modified from the raw set. Table \ref{tab:image_FPS} shows the results after the framerate of every video is set to 24fps. We observe that this has no real effect on image based approaches while only Waverep is significantly affected among the video-based methods. In Table \ref{tab:Resolution}, we present results after the resolution of video is mirrored from real videos. Here we observe a consistent drop in performance indicating that methods rely in part on difference in resolution to classify videos. In Table \ref{tab:Bitrates}, we show the effect of mirroring the bitrate form real videos. Here, we observe that some methods (e.g BFREE) are not affected while other like C-Forensics or WaveRep show a large drop in performance which indicate that some models rely more strongly on global image quality in their decision. Additionally we present in Table \ref{tab:duration} the effect on using video of different duration for real and generated samples. We observe We do not observe any significant change in performance here. While this is unsurprising for the image based methods, it indicates that video duration biases do not affect video-based detection methods. 
Overall we see that both image and video-based methods are mostly affected by biases in resolution and bitrate that are related to image quality. This is likely related to the training data as most commonly used datasets contains very large difference in resolutions between real and generated videos.

\begin{table}[ht]
    \centering
    \caption{Results of image (top) and video (bottom) based methods on the raw set.}

    \begin{subtable}{0.98\linewidth}
        \centering
        \caption{AUC / AP}
\begin{tabular}{lcccc}
            \hline
            \textbf{Method} & Grok & VEO3.1 & Sora2 & Global \\
            \hline
            B-Free\cite{guillaro2025bias} & 0.551 / 0.525 & 0.504 / 0.499 & 0.680 / 0.647 & 0.578 / 0.790 \\
            C2P-CLIP\cite{tan2025c2p} & 0.575 / 0.568 & 0.525 / 0.514 & 0.542 / 0.537 & 0.548 / 0.778 \\
            ForgeLens\cite{chen2025forgelens} & 0.781 / 0.790 & 0.702 / 0.710 & 0.622 / 0.664 & 0.702 / 0.723 \\
            D3-Im\cite{yang2025d}& 0.641 / 0.674 & 0.780 / 0.792 & 0.590 / 0.623 & 0.670/ 0.866 \\
            C-Forensics\cite{park2025community}& 0.793 / 0.789 & 0.790 / 0.789 & 0.783 / 0.787 & 0.789 / 0.913 \\
            \hline
            D3-Vid\cite{zheng2025d3vid}& 0.562 / 0.544 & 0.603 / 0.573 & 0.605 / 0.616 & 0.590 / 0.802 \\
            L3DE\cite{chang2025l3de} & 0.156 / 0.332 & 0.291 / 0.377 & 0.320 / 0.391 & 0.256 / 0.621 \\
            NSG-VD\cite{zhang2025NSGVD} & 0.654 / 0.598 & 0.689 / 0.638 & 0.723 / 0.698 & 0.689 / 0.844 \\
            WaveRep\cite{corvi2025waverep}& 0.976 / 0.976 & 0.934 / 0.946 & 0.736 / 0.762 & 0.882 / 0.913 \\
            STALL\cite{ben2026training} & 0.631 / 0.629 & 0.599 / 0.587 & 0.657 / 0.639 & 0.631 / 0.618 \\
            \hline
        \end{tabular}
    \end{subtable}
    
    \vspace{0.3cm}
    
    \begin{subtable}{0.98\linewidth}
        \centering
        \caption{Accuracy / Recall}
        \begin{tabular}{lcccc}
            \hline
            \textbf{Method} & Grok & VEO3.1 & Sora2 & Global \\
            \hline
            B-Free\cite{guillaro2025bias} & 0.546 / 0.493 & 0.516 / 0.309 & 0.597 / 0.309 & 0.563 / 0.425 \\
            C2P-CLIP\cite{tan2025c2p} & 0.511 / 0.998 & 0.509 / 0.993 & 0.511 / 0.998 & 0.510 / 0.996 \\
            ForgeLens\cite{chen2025forgelens}& 0.508 / 0.019 & 0.500 / 0.002 & 0.512 / 0.026 & 0.506 / 0.016 \\
            D3-Im\cite{yang2025d}&0.521 / 0.048 & 0.539 / 0.082 & 0.518 / 0.041 & 0.526 / 0.057 \\
            C-Forensics\cite{park2025community}& 0.505 / 0.010 & 0.519 / 0.039 & 0.520 / 0.041 & 0.515 / 0.030 \\
            \hline
            D3-Vid\cite{zheng2025d3vid}& 0.538 / 0.95 & 0.546 / 0.9652 & 0.517 / 0.909 & 0.534 / 0.941 \\
            L3DE\cite{chang2025l3de} & 0.361 / 0.037 & 0.389 / 0.094 & 0.403 / 0.127 & 0.384 / 0.084 \\
            NSG-VD\cite{zhang2025NSGVD} & 0.595 / 0.894 & 0.614 / 0.933 & 0.605 / 0.915 & 0.605 / 0.914 \\
            WaveRep\cite{corvi2025waverep}& 0.801 / 0.615 & 0.789 / 0.591 & 0.569 / 0.150 & 0.720 / 0.452 \\
            STALL\cite{ben2026training} & 0.573 / 0.316 & 0.564 / 0.316 & 0.590 / 0.316 & 0.576 / 0.316 \\

            \hline
        \end{tabular}
    \end{subtable}

    \vspace{0.3cm}

    \begin{subtable}{0.98\linewidth}
        \centering
        \caption{F1-score / ECE}
        \begin{tabular}{lcccc}
            \hline
            \textbf{Method} & Grok & VEO3.1 & Sora2 & Global \\
            \hline
            B-Free\cite{guillaro2025bias}& 0.521 / 0.130 & 0.390 / 0.106 & 0.532 / 0.115 & 0.557 / 0.114 \\
            C2P-CLIP\cite{tan2025c2p} & 0.671 / 0.015 & 0.669 / 0.015 & 0.671 / 0.015 & 0.858 / 0.015 \\
            ForgeLens\cite{chen2025forgelens}& 0.038 / 0.451 & 0.004 / 0.470 & 0.051 / 0.463 & 0.031 / 0.462 \\
            D3-Im\cite{yang2025d}& 0.090 / 0.443 & 0.152 / 0.414 & 0.079 / 0.450 & 0.108 / 0.435 \\
            C-Forensics\cite{park2025community}& 0.021 / 0.464 & 0.075 / 0.459 & 0.079 / 0.464 & 0.059 / 0.464 \\
            \hline
            D3-Vid\cite{zheng2025d3vid}& 0.673 / 0.284 & 0.680 / 0.298 & 0.653 / 0.316 & 0.843 / 0.118 \\
            L3DE\cite{chang2025l3de}& 0.055 / 0.626 & 0.133 / 0.591 & 0.169 / 0.574 & 0.141 / 0.748 \\
            NSG-VD\cite{zhang2025NSGVD}& 0.688 / 0.284 & 0.707 / 0.253 & 0.699 / 0.272 & 0.851 / 0.097 \\
            WaveRep\cite{corvi2025waverep}& 0.756 / 0.179 & 0.737 / 0.201 & 0.258 / 0.403 & 0.621 / 0.402 \\
            STALL\cite{ben2026training} & 0.425 / 0.199 & 0.420 / 0.196 & 0.435 / 0.208 & 0.427 / 0.201 \\

            \hline
        \end{tabular}
    \end{subtable}
    
    \vspace{0.3cm}
    
    \begin{subtable}{0.98\linewidth}
        \centering
        \caption{NLL}
        \begin{tabular}{lcccc}
            \hline
            \textbf{Method} & Grok & VEO3.1 & Sora2 & Global \\
            \hline
            B-Free\cite{guillaro2025bias} & 0.721 & 0.718 & 0.693 & 0.711 \\
            C2P-CLIP\cite{tan2025c2p} & 0.692 & 0.693 & 0.692 & 0.677 \\
            ForgeLens\cite{chen2025forgelens} & 1.679 & 1.899 & 1.923 & 1.839 \\
            D3-Im\cite{yang2025d} & 2.285 & 1.737 & 2.513 & 3.254 \\
            C-Forensics\cite{park2025community}& 2.278 & 2.745 & 2.801 & 4.162 \\
            \hline
            D3-Vid\cite{zheng2025d3vid}& 0.998 & 0.984 & 1.01 & 0.619 \\
            L3DE\cite{chang2025l3de} & 8.500 & 7.264 & 6.959 & 9.980 \\
            NSG-VD\cite{zhang2025NSGVD} & 0.877 & 0.848 & 0.867 & 0.555 \\
            WaveRep\cite{corvi2025waverep} & 0.475 & 0.734 & 1.956 & 1.558 \\
            STALL\cite{ben2026training} & 0.835 & 0.863 & 0.827 & 0.842 \\
            \hline
        \end{tabular}
    \end{subtable}

    \label{tab:raw}
\end{table}

\begin{table}[ht]
    \centering
    \caption{Results of image (top) and video (bottom) based methods on the Cleaned set.}

    \begin{subtable}{0.98\linewidth}
        \centering
        \caption{AUC / AP}
        \begin{tabular}{lcccc}
            \hline
            \textbf{Method} & Grok & VEO3.1 & Sora2 & Global \\
            \hline
            B-Free\cite{guillaro2025bias} & 0.427 / 0.473 & 0.359 / 0.401 & 0.465 / 0.467 & 0.417 / 0.704 \\
            C2P-CLIP\cite{tan2025c2p} & 0.409 / 0.426 &  0.419 / 0.430 & 0.402 / 0.424 & 0.410 / 0.684 \\
            ForgeLens\cite{chen2025forgelens} & 0.322 / 0.411 & 0.327 / 0.400 & 0.399 / 0.440 & 0.349 / 0.416\\ 
            D3-Im\cite{yang2025d}& 0.510 / 0.542 & 0.621 / 0.630 & 0.469 / 0.521 & 0.533 / 0.790\\
            C-Forensics\cite{park2025community}& 0.711 / 0.710 & 0.693 / 0.677 & 0.750 / 0.735 & 0.718 / 0.875 \\
            \hline
            D3-Vid\cite{zheng2025d3vid}& 0.506 / 0.480 & 0.523 / 0.490 & 0.543 / 0.532 & 0.524 / 0.749 \\
            L3DE\cite{chang2025l3de} & 0.663 / 0.630 & 0.701 / 0.661 & 0.631 / 0.601 & 0.665 / 0.835 \\
            NSG-VD\cite{zhang2025NSGVD} & 0.490 / 0.503 & 0.550 / 0.553 & 0.463 / 0.457 & 0.501 / 0.752 \\
            WaveRep\cite{corvi2025waverep} & 0.684 / 0.653 & 0.792 / 0.804 & 0.592 / 0.589 & 0.689 / 0.865 \\
                        STALL\cite{ben2026training} & 0.592 / 0.574 & 0.547 / 0.538 & 0.635 / 0.618 & 0.591 / 0.577 \\
            \hline
            
        \end{tabular}
    \end{subtable}
    
    \vspace{0.3cm}
    
    \begin{subtable}{0.98\linewidth}
        \centering
        \caption{Accuracy / Recall}
        \begin{tabular}{lcccc}
            \hline
            \textbf{Method} & Grok & VEO3.1 & Sora2 & Global \\
            \hline
            B-Free\cite{guillaro2025bias} & 0.460 / 0.248 & 0.359 / 0.170 & 0.465 / 0.302 & 0.450 / 0.229 \\
            C2P-CLIP\cite{tan2025c2p} & 0.510 / 0.996 & 0.502 / 0.978 &  0.504 / 0.985 & 0.505 / 0.986 \\
            ForgeLens\cite{chen2025forgelens}& 0.500 / 0.000 & 0.500 / 0.000& 0.500 / 0.000&0.500 / 0.000\\ 
            D3-Im\cite{yang2025d}& 0.508 / 0.020 & 0.511 / 0.026&  0.510 / 0.024 & 0.509 / 0.023\\
            C-Forensics\cite{park2025community}& 0.505 / 0.010 & 0.502 / 0.004 & 0.507 / 0.130 & 0.504 / 0.009 \\
            \hline
            D3-Vid\cite{zheng2025d3vid}& 0.540 / 0.974 & 0.547 / 0.972 & 0.504 / 0.998 & 0.530 / 0.999 \\
            L3DE\cite{chang2025l3de} & 0.588 / 0.867 & 0.602 / 0.896 & 0.575 / 0.841 & 0.588 / 0.868 \\
            NSG-VD\cite{zhang2025NSGVD} & 0.486 / 0.372 & 0.543 / 0.487 & 0.471 / 0.341 & 0.500 / 0.400 \\
            WaveRep\cite{corvi2025waverep} & 0.504 / 0.013 & 0.558 / 0.120 & 0.506 / 0.017 & 0.522 / 0.050 \\
                        STALL\cite{ben2026training} & 0.546 / 0.779 & 0.525 / 0.737 & 0.584 / 0.855 & 0.551 / 0.789 \\
            \hline
        \end{tabular}
    \end{subtable}

    \vspace{0.3cm}

    \begin{subtable}{0.98\linewidth}
        \centering
        \caption{F1-score / ECE}
        \begin{tabular}{lcccc}
            \hline
            \textbf{Method} & Grok & VEO3.1 & Sora2 & Global \\
            \hline
            B-Free\cite{guillaro2025bias} & 0.314 / 0.112 & 0.223 / 0.116 & 0.365 / 0.115 & 0.342 / 0.111 \\
            C2P-CLIP\cite{tan2025c2p} & 0.670 / 0.014 & 0.662 / 0.014 & 0.665 / 0.014 & 0.853 / 0.014\\
            ForgeLens\cite{chen2025forgelens}& 0.000 / 0.486 & 0.000 / 0.486& 0.000 / 0.488 & 0.000 / 0.486\\
            D3-Im\cite{yang2025d} & 0.038 / 0.469 & 0.051 / 0.459 & 0.047 / 0.459 &  0.045 / 0.465 \\ 
            C-Forensics\cite{park2025community}& 0.021 / 0.477 & 0.009 / 0.486 & 0.026 / 0.478 & 0.019 / 0.481\\
            \hline
             D3-Vid\cite{zheng2025d3vid}& 0.679 / 0.231 & 0.682 / 0.236 & 0.668 / 0.265 & 0.858 / 0.079 \\
            L3DE\cite{chang2025l3de} & 0.678 / 0.289 & 0.692 / 0.283 & 0.664 / 0.310 & 0.827 / 0.202 \\
            NSG-VD\cite{zhang2025NSGVD} & 0.420 / 0.085 & 0.516 / 0.026 & 0.392 / 0.100 & 0.522 / 0.277 \\
            WaveRep\cite{corvi2025waverep} & 0.026 / 0.478 & 0.213 / 0.423 & 0.034 / 0.479 & 0.095 / 0.694 \\
                        STALL\cite{ben2026training} & 0.408 / 0.214 & 0.397 / 0.209 & 0.429 / 0.201 & 0.411 / 0.208 \\
            \hline
        \end{tabular}
    \end{subtable}
    
    \vspace{0.3cm}
    
    \begin{subtable}{0.98\linewidth}
        \centering
        \caption{NLL}
        \begin{tabular}{lcccc}
            \hline
            \textbf{Method} & Grok & VEO3.1 & Sora2 & Global \\
            \hline
            B-Free\cite{guillaro2025bias} & 0.729 & 0.741 & 0.727 & 0.730 \\
            C2P-CLIP\cite{tan2025c2p} & 0.695 & 0.695 & 0.695 & 0.681\\
            ForgeLens\cite{chen2025forgelens}&2.307&2.267&2.321&2.298\\
            D3-Im\cite{yang2025d} & 2.828 & 2.451 &2.948 & 4.101\\
            C-Forensics\cite{park2025community}& 3.269 & 3.427 & 3.094 &4.891 \\
            \hline
            D3-Vid\cite{zheng2025d3vid}& 0.891 & 0.887 & 0.945 & 0.607 \\
            L3DE\cite{chang2025l3de} & 5.856 & 5.774 & 5.998 & 3.454 \\
            NSG-VD\cite{zhang2025NSGVD} & 0.710 & 0.693 & 0.715 & 0.733 \\
            WaveRep\cite{corvi2025waverep} & 2.505 & 1.912 & 2.792 & 3.595 \\
            STALL\cite{ben2026training} & 0.862 & 0.892 & 0.832 & 0.862 \\
            \hline
        \end{tabular}
    \end{subtable}

    \label{tab:image_cleaned}
\end{table}

\begin{table}[ht]
    \centering
    \caption{Results of image and video based methods on the Altered set.}

    \begin{subtable}{0.98\linewidth}
        \centering
        \caption{AUC / AP}
        \begin{tabular}{lcccc}
            \hline
            \textbf{Method} & Grok & VEO3.1 & Sora2 & Global \\
            \hline
            B-Free\cite{guillaro2025bias} & 0.461 / 0.451 & 0.450 / 0.438 & 0.484 / 0.464 & 0.468 / 0.721 \\
            C2P-CLIP\cite{tan2025c2p} & 0.468 / 0.451 &  0.450 / 0.438  & 0.484 / 0.464 & 0.467 / 0.704 \\
            ForgeLens\cite{chen2025forgelens} & 0.263 / 0.374 & 0.250 / 0.369 & 0.352 / 0.412 & 0.288 / 0.383\\ 
            D3-Im\cite{yang2025d}& 0.490 / 0.502 & 0.589 / 0.582 & 0.456 / 0.494 & 0.512 / 0.767\\
            C-Forensics\cite{park2025community}& 0.631 / 0.586 & 0.612 / 0.563 & 0.656 / 0.603 & 0.492 / 0.805 \\
            \hline
            D3-Vid\cite{zheng2025d3vid}& 0.370 / 0.404 & 0.400 / 0.421 & 0.424 / 0.437 & 0.398 / 0.678 \\
            L3DE\cite{chang2025l3de} & 0.652 / 0.624 & 0.676 / 0.640 & 0.613 / 0.585 & 0.647 / 0.827 \\
            NSG-VD\cite{zhang2025NSGVD} & 0.486 / 0.485 & 0.494 / 0.512 & 0.575 / 0.568 & 0.518 / 0.764 \\
            WaveRep\cite{corvi2025waverep} & 0.671 / 0.652 & 0.718 / 0.733 & 0.599 / 0.607 & 0.663 / 0.851 \\
            STALL\cite{ben2026training} & - / - & - / - & - / - & - / - \\
            \hline
        \end{tabular}
    \end{subtable}
    
    \vspace{0.3cm}
    
    \begin{subtable}{0.98\linewidth}
        \centering
        \caption{Accuracy / Recall}
        \begin{tabular}{lcccc}
            \hline
            \textbf{Method} & Grok & VEO3.1 & Sora2 & Global \\
            \hline
            B-Free\cite{guillaro2025bias} & 0.474 / 0.185 & 0.438 / 0.198 & 0.5220 / 0.361 & 0.473 / 0.229 \\
            C2P-CLIP\cite{tan2025c2p} & 0.513 / 0.939 & 0.502 / 0.917 &  0.513 / 0.939 & 0.509 / 0.932 \\
            ForgeLens\cite{chen2025forgelens}& 0.499 / 0.002 & 0.499 / 0.002& 0.500 / 0.000&0.499 / 0.001\\ 
            D3-Im\cite{yang2025d}& 0.493 / 0.015 & 0.508 / 0.044&  0.498 / 0.024 & 0.499 / 0.028\\
            C-Forensics\cite{park2025community}& 0.493 / 0.017 & 0.489 / 0.008 & 0.493 / 0.017 & 0.253 / 0.014 \\
            \hline
            D3-Vid\cite{zheng2025d3vid}& 0.503 / 0.996 & 0.502 / 1.000 & 0.500 / 1.000 & 0.502 / 1.000 \\
            L3DE\cite{chang2025l3de} & 0.586 / 0.844 & 0.599 / 0.870 & 0.569 / 0.809 & 0.584 / 0.841 \\
            NSG-VD\cite{zhang2025NSGVD} & 0.483 / 0.183 & 0.509 / 0.235 & 0.548 / 0.313 & 0.513 / 0.243 \\
            WaveRep\cite{corvi2025waverep} & 0.506 / 0.013 & 0.538 / 0.076 & 0.515 / 0.030 & 0.520 / 0.040 \\
                        STALL\cite{ben2026training} & - / - & - / - & - / - & - / - \\

            \hline
        \end{tabular}
    \end{subtable}

    \vspace{0.3cm}

    \begin{subtable}{0.98\linewidth}
        \centering
        \caption{F1-score / ECE}
        \begin{tabular}{lcccc}
            \hline
            \textbf{Method} & Grok & VEO3.1 & Sora2 & Global \\
            \hline
            B-Free\cite{guillaro2025bias} & 0.260 / 0.102 & 0.260 / 0.116 & 0.429 / 0.117 & 0.346 / 0.111 \\
            C2P-CLIP\cite{tan2025c2p} & 0.658 / 0.011 & 0.648 / 0.011 & 0.658 / 0.011 & 0.833 / 0.011\\
            ForgeLens\cite{chen2025forgelens}& 0.004 / 0.487 & 0.000 / 0.488& 0.004 / 0.487 & 0.003 / 0.487\\
            D3-Im\cite{yang2025d} & 0.029 / 0.468 & 0.081 / 0.442 & 0.045 / 0.460 &  0.053 / 0.456 \\ 
            C-Forensics\cite{park2025community}& 0.033 / 0.461 & 0.017 / 0.476 & 0.033 / 0.466 & 0.028 / 0.468\\
            \hline
            D3-Vid\cite{zheng2025d3vid}& 0.667 / 0.274 & 0.668 / 0.243 & 0.667 / 0.243 & 0.857 / 0.221 \\
            L3DE\cite{chang2025l3de} & 0.671 / 0.284 & 0.684 / 0.276 & 0.652 / 0.302 & 0.814 / 0.213 \\
            NSG-VD\cite{zhang2025NSGVD} & 0.261 / 0.088 & 0.323 / 0.062 & 0.409 / 0.052 & 0.370 / 0.307 \\
            WaveRep\cite{corvi2025waverep} & 0.026 / 0.474 & 0.141 / 0.441 & 0.059 / 0.472 & 0.077 / 0.700 \\
                        STALL\cite{ben2026training} & - / - & - / - & - / - & - / - \\

            \hline
        \end{tabular}
    \end{subtable}
    
    \vspace{0.3cm}
    
    \begin{subtable}{0.98\linewidth}
        \centering
        \caption{NLL}
        \begin{tabular}{lcccc}
            \hline
            \textbf{Method} & Grok & VEO3.1 & Sora2 & Global \\
            \hline
            B-Free\cite{guillaro2025bias} & 0.720 & 0.732 & 0.721 & 0.723 \\
            C2P-CLIP\cite{tan2025c2p} & 0.694 & 0.694 & 0.694 & 0.683\\
            ForgeLens\cite{chen2025forgelens} & 2.267 & 2.269 & 2.201 & 2.246\\
            D3-Im\cite{yang2025d} & 2.597 & 2.236 &2.693 & 3.728\\
            C-Forensics\cite{park2025community}& 2.714 & 2.826 & 2.577 &4.030 \\
            \hline
            D3-Vid\cite{zheng2025d3vid}& 0.986 & 0.946 & 0.940 & 0.722 \\
            L3DE\cite{chang2025l3de} & 6.291 & 6.090 & 6.397 & 3.837 \\
            NSG-VD\cite{zhang2025NSGVD} & 0.710 & 0.707 & 0.690 & 0.761 \\
            WaveRep\cite{corvi2025waverep} & 2.385 & 2.111 & 2.587 & 3.535 \\
                        STALL\cite{ben2026training} &  - &  - &  - &  - \\

            \hline
        \end{tabular}
    \end{subtable}

    \label{tab:image_altered}
\end{table}

    \begin{table}[ht]
        \centering
        \caption{AUC results of image (top) and video (bottom) based methods on GenVideo.}
        \makebox[\textwidth][c]{
        \begin{tabular}{lccccccccccc}
            \hline
            \textbf{Method} & Crafter & Gen2 & HotShot & Lavie & ModelScope & MoonValley & MorphStudio & Show-1 & Sora & WildScrape & Global \\
            \hline
            B-Free\cite{guillaro2025bias} & 1.000  &  0.999 &  0.994 &  0.999 & 0.994  &  1.000 &  0.999 &  0.993 &  0.811 &  0.905 &  0.987 \\
            C2P-CLIP\cite{tan2025c2p}  & 0.884  &  0.982 &  0.857 &  0.874 & 0.887  &  0.973 & 0.921  & 0.837  & 0.887  &  0.843 &  0.8974 \\
            ForgeLens\cite{chen2025forgelens}  & 0.971  &  0.845 &  0.288 & 0.778 & 0.724  &  0.348 &  0.593 &  0.682 & 0.712  &  0.590 &  0.701 \\
            D3-Im\cite{yang2025d} & 0.991  &  0.972 &  0.896 &  0.975 & 0.942  &  0.985 &  0.974 &  0.943 & 0.915  &  0.869 &  0.954 \\
            C-Forensics\cite{park2025community} & 0.999  &  0.994 &  0.986 &  0.995 & 0.993  &  0.998 &  0.998 &  0.996 & 0.796  &  0.798 &  0.973 \\
            \hline
            D3-Vid\cite{zheng2025d3vid} & 0.982  &  0.991 &  0.974 &  0.970 & 0.958  &  0.990 &  0.975 &  0.984 & 0.977  &  0.919&  0.972 \\
            L3DE\cite{chang2025l3de}  & 0.406  &  0.578 &  0.414 &  0.484 & 0.602  &  0.311 &  0.527 &  0.715 & 0.3409  &  0.509 &  0.502 \\
            NSG-VD\cite{zhang2025NSGVD} & 0.986  &  0.963 &  0.900 &  0.914 & 0.875  &  0.997 &  0.958 &  0.913 & 0.956  &  0.905 &  0.938 \\
            WaveRep\cite{corvi2025waverep} & 1.000  &  1.000 &  0.942 &  0.993 & 0.997  &  1.000 &  0.999 &  0.994 & 0.836  &  0.908 &  0.982 \\
            STALL\cite{ben2026training} & 0.756  &  0.787 &  0.703 &  0.763 & 0.662  &  0.615 &  0.718 &  0.729 & 0.707  & 0.601 &  0.704 \\
            \hline
        \end{tabular}}
    \label{tab:genvideo_AUC}
\end{table}
    
    \vspace{0.3cm}
    
    \begin{table}[ht]
        \centering
        \caption{AP results of image (top) and video (bottom) based methods on GenVideo.}
        \makebox[\textwidth][c]{
        \begin{tabular}{lccccccccccc}
            \hline
            \textbf{Method} & Crafter & Gen2 & HotShot & Lavie & ModelScope & MoonValley & MorphStudio & Show-1 & Sora & WildScrape & Global \\
            \hline
            B-Free\cite{guillaro2025bias} & 0.998  &  0.995 &  0.961 &  0.993 & 0.954  &  0.998 &  0.993 &  0.968 & 0.077  &  0.801 &  0.990 \\
            C2P-CLIP\cite{tan2025c2p}  & 0.718  &  0.923 &  0.466 &  0.653 & 0.609  &  0.821 &  0.672 &  0.431 & 0.091  &  0.493 &  0.9031 \\
            ForgeLens\cite{chen2025forgelens}  & 0.923  &  0.649 &  0.109 &  0.715 & 0.546  &  0.091 &  0.342 &  0.399 & 0.026  &  0.341 &  0.530 \\
            D3-Im\cite{yang2025d} & 0.976  &  0.916 &  0.657 &  0.925 & 0.780  &  0.949 &  0.892 &  0.773 & 0.211  &  0.645 &  0.960 \\
            C-Forensics\cite{park2025community} & 0.999  &  0.981 &  0.953 &  0.987 & 0.965  &  0.991 &  0.994 &  0.980 & 0.063  &  0.662 &  0.981 \\
            \hline
            D3-Vid\cite{zheng2025d3vid} & 0.873  &  0.909 &  0.691 &  0.806 & 0.627  &  0.837 &  0.679 &  0.752 & 0.192  &  0.559 &  0.962 \\
            L3DE\cite{chang2025l3de}  & 0.097  &  0.147 &  0.052 &  0.115 & 0.092  &  0.039 &  0.068 &  0.116 & 0.004  &  0.104 &  0.463 \\
            NSG-VD\cite{zhang2025NSGVD} & 0.904  &  0.809 &  0.435 &  0.604 & 0.507  &  0.936 &  0.625 &  0.388 & 0.221  &  0.638 &  0.931 \\
            WaveRep\cite{corvi2025waverep} & 0.999  &  1.000 &  0.755 &  0.982 & 0.973  &  0.998 &  0.995 &  0.966 & 0.418  &  0.804 &  0.987 \\
            STALL\cite{ben2026training} & 0.752  &  0.800 &  0.674 &  0.741 & 0.642  &  0.646 &  0.716 &  0.698 & 0.704  &  0.554 &  0.693 \\

            \hline
        \end{tabular}}
    \label{tab:genvideo_AP}
\end{table}

    \vspace{0.3cm}

    \begin{table}[ht]
        \centering
        \caption{Balanced accuracy results of image (top) and video (bottom) based methods on GenVideo.}
        \makebox[\textwidth][c]{
        \begin{tabular}{lccccccccccc}
            \hline
            \textbf{Method} & Crafter & Gen2 & HotShot & Lavie & ModelScope & MoonValley & MorphStudio & Show-1 & Sora & WildScrape & Global \\
            \hline
            B-Free\cite{guillaro2025bias} & 0.970  &  0.967 &  0.955 &  0.967 & 0.953  &  0.967 &  0.968 &  0.964 & 0.754  &  0.854 &  0.948 \\
            C2P-CLIP\cite{tan2025c2p}  & 0.514  &  0.515 &  0.514 &  0.514 & 0.515  &  0.515 &  0.514 &  0.515 & 0.515  &  0.513 &  0.514 \\
            ForgeLens\cite{chen2025forgelens}  & 0.879  &  0.632 &  0.527 &  0.766 & 0.656  &  0.515 &  0.597 &  0.598 & 0.516  &  0.579 &  0.667 \\
            D3-Im\cite{yang2025d} & 0.968  &  0.904 &  0.775 &  0.915 & 0.836  &  0.957 &  0.909 &  0.830 & 0.715  &  0.743 &  0.880 \\
            C-Forensics\cite{park2025community} & 0.992  &  0.968 &  0.954 &  0.978 & 0.962  &  0.986 &  0.989 &  0.972 & 0.564  &  0.774 &  0.952 \\
            \hline
            D3-Vid\cite{zheng2025d3vid} & 0.935  &  0.951 &  0.924 &  0.913 & 0.901  &  0.947 &  0.931 &  0.943 & 0.935  &  0.852 &  0.923 \\
            L3DE\cite{chang2025l3de}  & 0.422  &  0.551 &  0.428 &  0.478 & 0.571  &  0.349 &  0.493 &  0.673 & 0.392  &  0.505 &  0.486 \\
            NSG-VD\cite{zhang2025NSGVD} & 0.929  &  0.898 &  0.809 &  0.827 & 0.808  &  0.938 &  0.889 &  0.833 & 0.884  &  0.830 &  0.866 \\
            WaveRep\cite{corvi2025waverep} & 0.989  &  0.989 &  0.834 &  0.973 & 0.975  &  0.989 &  0.986 &  0.968 & 0.766  &  0.858 &  0.933 \\
            STALL\cite{ben2026training} & 0.530  &  0.534 &  0.519 &  0.528 & 0.516  &  0.531 &  0.526 &  0.520 & 0.518  & 0.495  &  0.522 \\

            \hline
        \end{tabular}}
                \label{tab:genvideo_ACC}

\end{table}
    
    \vspace{0.3cm}
    
        \begin{table}[ht]
        \centering

        \caption{Recall results of image (top) and video (bottom) based methods on GenVideo.}
        \makebox[\textwidth][c]{
        \begin{tabular}{lccccccccccc}
            \hline
            \textbf{Method} & Crafter & Gen2 & HotShot & Lavie & ModelScope & MoonValley & MorphStudio & Show-1 & Sora & WildScrape & Global \\
            \hline
            B-Free\cite{guillaro2025bias} & 0.999  &  0.999 &  0.974 &  0.992 & 0.981  &  1.000 &  0.996 &  0.980 & 0.750  &  0.827 &  0.969 \\
            C2P-CLIP\cite{tan2025c2p}  & 0.997  &  1.000 &  0.997 &  0.999 & 1.000  &  1.000 &  0.999 &  1.000 & 1.000  &  0.996 &  0.999 \\
            ForgeLens\cite{chen2025forgelens}  & 0.759  &  0.267 &  0.057 &  0.535 & 0.316  &  0.034 &  0.197 &  0.198 & 0.035  &  0.162 &  0.336 \\
            D3-Im\cite{yang2025d} & 0.953  &  0.824 &  0.565 &  0.845 & 0.688  &  0.931 &  0.834 &  0.677 & 0.446  &  0.502 &  0.776 \\
            C-Forensics\cite{park2025community} & 0.998  &  0.949 &  0.922 &  0.969 & 0.937  &  0.985 &  0.992 &  0.958 & 0.142  &  0.562 &  0.917 \\
            \hline
            D3-Vid\cite{zheng2025d3vid} & 0.964  &  0.996 &  0.943 &  0.92 & 0.896  &  0.989 &  0.956 &  0.981 & 0.964  &  0.798 &  0.940 \\
            L3DE\cite{chang2025l3de}  & 0.239  &  0.496 &  0.250 &  0.351 & 0.536  &  0.093 &  0.380 &  0.740 & 0.179  &  0.405 &  0.383 \\
            NSG-VD\cite{zhang2025NSGVD} & 0.982  &  0.920 &  0.743 &  0.779 & 0.741  &  1.000 &  0.903 &  0.791& 0.893  &  0.786 &  0.858 \\
            WaveRep\cite{corvi2025waverep} & 0.999  &  0.999 &  0.689 &  0.967 & 0.971  &  0.998 &  0.993 &  0.957 & 0.554  &  0.736 &  0.931 \\
            STALL\cite{ben2026training} & 0.069  &  0.069 &  0.063 &  0.069 & 0.063  &  0.065 &  0.063 &  0.063 & 0.071  &  0.065 &  0.066 \\

            \hline
        \end{tabular}}
                        \label{tab:genvideo_RECALL}

\end{table}

        \vspace{0.3cm}
    
        \begin{table}[ht]
        \centering

        \caption{F1-score results of image (top) and video (bottom) based methods on GenVideo.}
        \makebox[\textwidth][c]{
        \begin{tabular}{lccccccccccc}
            \hline
            \textbf{Method} & Crafter & Gen2 & HotShot & Lavie & ModelScope & MoonValley & MorphStudio & Show-1 & Sora & WildScrape & Global \\
            \hline
            B-Free\cite{guillaro2025bias} & 0.828  &  0.808 &  0.674 &  0.826 & 0.639  &  0.657 &  0.696 &  0.720 & 0.033  &  0.531 &  0.944 \\
            C2P-CLIP\cite{tan2025c2p}  & 0.223  &  0.222 &  0.126 &  0.224 & 0.126  &  0.114 &  0.126 &  0.126 & 0.011  &  0.159 &  0.638 \\
            ForgeLens\cite{chen2025forgelens}  & 0.854  &  0.415 &  0.104 &  0.689 & 0.467  &  0.062 &  0.319 &  0.321 & 0.048  &  0.272 &  0.493 \\
            D3-Im\cite{yang2025d} & 0.921  &  0.850 &  0.632 &  0.863 & 0.719  &  0.853 &  0.810 &  0.712 & 0.210  &  0.600 &  0.865 \\
            C-Forensics\cite{park2025community} &0.955  &  0.929 &  0.875 &  0.940 & 0.883  &  0.899 &  0.912 &  0.894 & 0.082  &  0.661 &  0.949 \\
            \hline
            D3-Vid\cite{zheng2025d3vid} & 0.732  &  0.745 &  0.575 &  0.711 & 0.554  &  0.568 &  0.581 &  0.592 & 0.103  &  0.565 &  0.917 \\
            L3DE\cite{chang2025l3de}  & 0.119  &  0.229 &  0.073 &  0.169 & 0.151  &  0.025 &  0.109 &  0.202 & 0.005  &  0.144 &  0.417 \\
            NSG-VD\cite{zhang2025NSGVD} & 0.686  &  0.654 &  0.424 &  0.585 & 0.423  &  0.503 &  0.492 &  0.445 & 0.075  &  0.501 &  0.857 \\
            WaveRep\cite{corvi2025waverep} & 0.931  &  0.931 &  0.696 &  0.916 & 0.859  &  0.859 &  0.869 &  0.851 & 0.213  &  0.752 &  0.953 \\
            STALL\cite{ben2026training} & 0.127  &  0.129 &  0.115 &  0.127 & 0.115  &  0.123 &  0.117 &  0.116 & 0.129  &  0.114 &  0.121 \\

            \hline
        \end{tabular}}
                        \label{tab:genvideo_F1}

\end{table}

    \vspace{0.3cm}
    
        \begin{table}[ht]
        \centering

        \caption{ECE results of image (top) and video (bottom) based methods on GenVideo.}
        \makebox[\textwidth][c]{
        \begin{tabular}{lccccccccccc}
            \hline
            \textbf{Method} & Crafter & Gen2 & HotShot & Lavie & ModelScope & MoonValley & MorphStudio & Show-1 & Sora & WildScrape & Global \\
            \hline
            B-Free\cite{guillaro2025bias} & 0.433  &  0.422 &  0.393 &  0.417 & 0.381  &  0.431 &  0.429 &  0.396 & 0.230  &  0.328 &  0.403 \\
            C2P-CLIP\cite{tan2025c2p}  & 0.017  &  0.024 &  0.015 &  0.016 & 0.016  &  0.018 &  0.016 &  0.015 & 0.015  &  0.015 &  0.017 \\
            ForgeLens\cite{chen2025forgelens}  & 0.118  &  0.341 &  0.451 &  0.225 & 0.316  &  0.457 &  0.377 &  0.371 & 0.442  &  0.396 &  0.313 \\
            D3-Im\cite{yang2025d} & 0.019  &  0.072 &  0.189 &  0.063 & 0.137  &  0.021 &  0.063 &  0.139 & 0.233  &  0.213 &  0.090 \\
            C-Forensics\cite{park2025community} & 0.027  &  0.027 &  0.019 &  0.022 & 0.045  &  0.025 &  0.025 &  0.026 & 0.352  &  0.194 &  0.030 \\
            \hline
            D3-Vid\cite{zheng2025d3vid} & 0.088  &  0.094 &  0.099 &  0.083 & 0.096  &  0.104 &  0.099 &  0.101 & 0.114  &  0.087 &  0.022 \\
            L3DE\cite{chang2025l3de}  & 0.426  &  0.393 &  0.410 &  0.411 & 0.390  &  0.418 &  0.398 &  0.379 & 0.395  &  0.403 &  0.469 \\
            NSG-VD\cite{zhang2025NSGVD} & 0.147  &  0.141 &  0.150 &  0.125 & 0.150  &  0.167 &  0.159 &  0.152 & 0.180  &  0.145 &  0.029 \\
            WaveRep\cite{corvi2025waverep} & 0.042  &  0.042 &  0.024 &  0.037 & 0.042  &  0.045 &  0.044 &  0.041 & 0.045  &  0.025 &  0.012 \\
            STALL\cite{ben2026training} & 0.375  &  0.381 &  0.368 &  0.375 & 0.356  &  0.350 &  0.368 &  0.369 & 0.389  &  0.362 &  0.369 \\

            \hline
        \end{tabular}}
                        \label{tab:genvideo_ECE}

\end{table}

    \vspace{0.3cm}
    
       \begin{table}[ht]
        \centering

        \caption{NLL results of image (top) and video (bottom) based methods on GenVideo.}
        \makebox[\textwidth][c]{
        \begin{tabular}{lccccccccccc}
            \hline
            \textbf{Method} & Crafter & Gen2 & HotShot & Lavie & ModelScope & MoonValley & MorphStudio & Show-1 & Sora & WildScrape & Global \\
            \hline
            B-Free\cite{guillaro2025bias} & 0.612  &  0.618 &  0.627 &  0.617 & 0.631  &  0.612 &  0.617 &  0.625 & 0.682  &  0.653 &  0.629 \\
            C2P-CLIP\cite{tan2025c2p}  & 0.709  &  0.708 &  0.713 &  0.709 & 0.713  &  0.713 &  0.713 &  0.713 & 0.716  &  0.712 &  0.687 \\
            ForgeLens\cite{chen2025forgelens}  & 0.105  &  0.287 &  0.309 &  0.239 & 0.177  &  0.274 &  0.231 &  0.212 & 0.047  &  0.298 &  0.219 \\
            D3-Im\cite{yang2025d} & 0.081  &  0.117 &  0.149 &  0.112 & 0.121  &  0.082 &  0.095 &  0.120 & 0.081  &  0.189 &  0.309 \\
            C-Forensics\cite{park2025community} & 0.060  &  0.074 &  0.076 &  0.070 & 0.074  &  0.066 &  0.065 &  0.069 & 0.080  &  0.237 &  0.182 \\
            \hline
            D3-Vid\cite{zheng2025d3vid} & 0.196  &  0.187 &  0.203 &  0.209 & 0.214  &  0.196 &  0.202 &  0.198 & 0.204  &  0.249 &  0.216 \\
            L3DE\cite{chang2025l3de}  & 4.696  &  4.209 &  4.358 &  4.445 & 4.086  &  4.489 &  4.182 &  3.907 & 4.072  &  4.305 &  5.473 \\
            NSG-VD\cite{zhang2025NSGVD} & 0.277  &  0.290 &  0.316 &  0.320 & 0.324  &  0.286 &  0.295 &  0.310 & 0.299  &  0.319 &  0.326 \\
            WaveRep\cite{corvi2025waverep} & 0.062  &  0.062 &  0.122 &  0.074 & 0.071  &  0.067 &  0.067 &  0.073 & 0.079  &  0.148 &  0.139 \\
            STALL\cite{ben2026training} & 1.173  &  1.167 &  1.190 &  1.176 & 1.198  &  1.199 & 0.179  &  0.184 & 0.151  &  1.234 &  1.185 \\

            \hline
        \end{tabular}}
                        \label{tab:genvideo_NLL}

\end{table}

\begin{table}[ht]
    \centering
    \caption{Results of image (top) and video (bottom) based methods on CCDF raw set with FPS set to 24.}

    \begin{subtable}{0.98\linewidth}
        \centering
        \caption{AUC / AP}
        \begin{tabular}{lcccc}
            \hline
            \textbf{Method} & Grok & VEO3.1 & Sora2 & Global \\
            \hline
            B-Free\cite{guillaro2025bias} & 0.464 / 0.458 & 0.504 / 0.495 & 0.624 / 0.589 & 0.531 / 0.758\\
            C2P-CLIP\cite{tan2025c2p} & 0.518 / 0.513 & 0.521 / 0.501 & 0.592 / 0.564 & 0.544 / 0.768 \\
            ForgeLens\cite{chen2025forgelens} & 0.455 / 0.492 & 0.550 / 0.553 & 0.438 / 0.481 & 0.481 / 0.508 \\
            D3-Im\cite{yang2025d}& 0.595 / 0.631 & 0.738 / 0.753 & 0.561 / 0.600 & 0.632 / 0.847 \\
            C-Forensics\cite{park2025community}& 0.805 / 0.799 & 0.784 / 0.799 & 0.770 / 0.775 & 0.786 / 0.912 \\
            \hline
            D3-Vid\cite{zheng2025d3vid}& 0.587 / 0.548 & 0.604 / 0.563 & 0.584 / 0.600 & 0.592 / 0.799 \\
            L3DE\cite{chang2025l3de} & 0.209 / 0.346 & 0.302 / 0.380 & 0.331 / 0.394 & 0.281 / 0.630 \\
            NSG-VD\cite{zhang2025NSGVD} & 0.617 / 0.573 & 0.653 / 0.612 & 0.694 / 0.678 & 0.655 / 0.831 \\
            WaveRep\cite{corvi2025waverep}& 0.876 / 0.884 & 0.810 / 0.843 & 0.533 / 0.560 & 0.740 / 0.904 \\
            STALL\cite{ben2026training} & 0.628 / 0.614 & 0.594 / 0.582 & 0.655 / 0.636 & 0.626 / 0.611 \\

            \hline
        \end{tabular}
    \end{subtable}
    
    \vspace{0.3cm}
    
    \begin{subtable}{0.98\linewidth}
        \centering
        \caption{Accuracy / Recall}
        \begin{tabular}{lcccc}
            \hline
            \textbf{Method} & Grok & VEO3.1 & Sora2 & Global \\
            \hline
            B-Free\cite{guillaro2025bias} & 0.487 / 0.348 & 0.518 / 0.333  & 0.563 / 0.393 & 0.529 / 0.379 \\
            C2P-CLIP\cite{tan2025c2p} & 0.508 / 0.991 & 0.509 / 0.993 & 0.513 / 1.0 & 0.510 / 0.995 \\
            ForgeLens\cite{chen2025forgelens}& 0.499 / 0.000 & 0.499 / 0.000 & 0.500 / 0.002 & 0.499 / 0.001 \\
            D3-Im\cite{yang2025d}& 0.515 / 0.034 & 0.520 / 0.045 & 0.515 / 0.034 & 0.517 / 0.038 \\
            C-Forensics\cite{park2025community}& 0.505 / 0.010 & 0.513 / 0.026 & 0.511 / 0.023 & 0.510 / 0.020 \\
            \hline
            D3-Vid\cite{zheng2025d3vid}& 0.546 / 0.933 & 0.560 / 0.961 & 0.524 / 0.889 & 0.543 / 0.928 \\
            L3DE\cite{chang2025l3de} & 0.372 / 0.054 & 0.395 / 0.100 & 0.411 / 0.133 & 0.392 / 0.096 \\
            NSG-VD\cite{zhang2025NSGVD} & 0.573 / 0.861 & 0.600 / 0.915 & 0.587 / 0.889 & 0.587 / 0.888 \\
            WaveRep\cite{corvi2025waverep}& 0.626 / 0.265 & 0.638 / 0.289 & 0.505 / 0.024 & 0.590 / 0.193 \\
            STALL\cite{ben2026training} & 0.551 / 0.809 & 0.542 / 0.882 & 0.579 / 0.865 & 0.558 / 0.823 \\

            \hline
        \end{tabular}
    \end{subtable}

    \vspace{0.3cm}

    \begin{subtable}{0.98\linewidth}
        \centering
        \caption{F1-score / ECE}
        \begin{tabular}{lcccc}
            \hline
            \textbf{Method} & Grok & VEO3.1 & Sora2 & Global \\
            \hline
            B-Free\cite{guillaro2025bias}& 0.404 / 0.124 & 0.409 / 0.109 & 0.474 / 0.104 & 0.510 / 0.113 \\
            C2P-CLIP\cite{tan2025c2p} & 0.668 / 0.015 & 0.669 / 0.15 & 0.672 / 0.016 & 0.858 / 0.015 \\
            ForgeLens\cite{chen2025forgelens}& 0.000 / 0.482 & 0.000 / 0.483 & 0.004 / 0.481 & 0.001 / 0.481 \\
            D3-Im\cite{yang2025d}& 0.066 / 0.453 & 0.086 / 0.429 & 0.066 / 0.457 & 0.073 / 0.446 \\
            C-Forensics\cite{park2025community}& 0.021 / 0.470 & 0.050 / 0.463 & 0.023 / 0.469 & 0.039 / 0.467 \\
            \hline
            D3-Vid\cite{zheng2025d3vid}& 0.672 / 0.292 & 0.686 / 0.294 & 0.651 / 0.307 & 0.840 / 0.116 \\
            L3DE\cite{chang2025l3de}& 0.080 / 0.610 & 0.142 / 0.580 & 0.184 / 0.592 & 0.160 / 0.737 \\
            NSG-VD\cite{zhang2025NSGVD}& 0.668 / 0.260 & 0.696 / 0.255 & 0.683 / 0.261 & 0.835 / 0.105 \\
            WaveRep\cite{corvi2025waverep}& 0.415 / 0.340 & 0.444 / 0.338 & 0.046 / 0.474 & 0.322 / 0.584 \\
            STALL\cite{ben2026training} & 0.395 / 0.213 & 0.391 / 0.208 & 0.411 / 0.210 & 0.399 / 0.210 \\

            \hline
        \end{tabular}
    \end{subtable}
    
    \vspace{0.3cm}
    
    \begin{subtable}{0.98\linewidth}
        \centering
        \caption{NLL}
        \begin{tabular}{lcccc}
            \hline
            \textbf{Method} & Grok & VEO3.1 & Sora2 & Global \\
            \hline
            B-Free\cite{guillaro2025bias} & 0.731 & 0.720 & 0.701 & 0.718 \\
            C2P-CLIP\cite{tan2025c2p} & 0.693 & 0.693 & 0.692 & 0.677 \\
            ForgeLens\cite{chen2025forgelens} & 2.191 & 2.131 & 2.206 & 2.176 \\
            D3-Im\cite{yang2025d} & 2.478 & 1.934 & 2.610 & 3.498 \\
            C-Forensics\cite{park2025community}& 2.731 & 2.805 & 2.907 & 4.218 \\
            \hline
            D3-Vid\cite{zheng2025d3vid}& 0.986 & 0.979 & 1.027 & 0.6263 \\
            L3DE\cite{chang2025l3de} & 8.067 & 7.123 & 6.838 & 9.6269 \\
            NSG-VD\cite{zhang2025NSGVD} & 0.909 & 0.877 & 0.878 & 0.5846 \\
            WaveRep\cite{corvi2025waverep} & 1.265 & 1.494 & 2.817 & 2.7728 \\
            STALL\cite{ben2026training} & 0.846 & 0.869 & 0.831 & 0.849 \\

            \hline
        \end{tabular}
    \end{subtable}

    \label{tab:image_FPS}
\end{table}

\begin{table}[ht]
    \centering
    \caption{Results of image (top) and video (bottom) based methods on CCDF raw set with resolution mirrored from real videos.}

    \begin{subtable}{0.98\linewidth}
        \centering
        \caption{AUC / AP}
        \begin{tabular}{lcccc}
            \hline
            \textbf{Method} & Grok & VEO3.1 & Sora2 & Global \\
            \hline
            B-Free\cite{guillaro2025bias} & 0.374 / 0.495 & 0.340 / 0.394 & 0.475 / 0.496 & 0.396 / 0.716 \\
            C2P-CLIP\cite{tan2025c2p} & 0.459 / 0.466 & 0.450 / 0.453 & 0.434 / 0.449 & 0.448 / 0.713 \\
            ForgeLens\cite{chen2025forgelens} & 0.383 / 0.464 & 0.354 / 0.421 & 0.463 / 0.483 & 0.400 / 0.454 \\
            D3-Im\cite{yang2025d}& 0.537 / 0.581 & 0.657 / 0.679 & 0.487 / 0.538 & 0.560 / 0.811 \\
            C-Forensics\cite{park2025community}& 0.748 / 0.887 & 0.695 / 0.685 & 0.772 / 0.760 & 0.738 / 0.887 \\
            \hline
            D3-Vid\cite{zheng2025d3vid} & 0.533 / 0.523 & 0.560 / 0.532 & 0.542 / 0.552 & 0.545 / 0.773 \\
            L3DE\cite{chang2025l3de} & 0.267 / 0.366 & 0.267 / 0.366 & 0.370 / 0.412 & 0.301 / 0.638 \\
            NSG-VD\cite{zhang2025NSGVD} & 0.533 / 0.522 & 0.518 / 0.504 & 0.594 / 0.565 & 0.548 / 0.771 \\
            WaveRep\cite{corvi2025waverep} & 0.830 / 0.808 & 0.867 / 0.858 & 0.603 / 0.582 & 0.767 / 0.901 \\
            STALL\cite{ben2026training} & 0.612 / 0.597 & 0.560 / 0.554 & 0.631 / 0.622 & 0.601 / 0.591 \\

            \hline
        \end{tabular}
    \end{subtable}
    
    \vspace{0.3cm}
    
    \begin{subtable}{0.98\linewidth}
        \centering
        \caption{Accuracy / Recall}
        \begin{tabular}{lcccc}
            \hline
            \textbf{Method} & Grok & VEO3.1 & Sora2 & Global \\
            \hline
            B-Free\cite{guillaro2025bias} & 0.463 / 0.207 & 0.408 / 0.170 & 0.484 / 0.250 & 0.454 / 0.189 \\
            C2P-CLIP\cite{tan2025c2p} & 0.505 / 0.987 & 0.503 / 0.983 & 0.504 / 0.985 & 0.504 / 0.985 \\
            ForgeLens\cite{chen2025forgelens}& 0.499 / 0.000 & 0.499 / 0.000 & 0.500 / 0.002 & 0.499 / 0.001 \\
            D3-Im\cite{yang2025d}& 0.513/ 0.028 & 0.518 / 0.039 & 0.514 / 0.030 & 0.515 / 0.032 \\
            C-Forensics\cite{park2025community}& 0.513 / 0.026 & 0.501 / 0.002 & 0.505 / 0.010 & 0.506 / 0.013 \\
            \hline
            D3-Vid\cite{zheng2025d3vid} & 0.549 / 0.954 & 0.554 / 0.965 & 0.529 / 0.915 & 0.544 / 0.945 \\
            L3DE\cite{chang2025l3de} & 0.392 / 0.098 & 0.382 / 0.076 & 0.426 / 0.165 & 0.400 / 0.113 \\
            NSG-VD\cite{zhang2025NSGVD} & 0.523 / 0.737 & 0.521 / 0.733 & 0.559 / 0.809 & 0.534 / 0.759 \\
            WaveRep\cite{corvi2025waverep} & 0.551 / 0.113 & 0.564 / 0.139 & 0.506 / 0.024 & 0.541 / 0.092 \\
            STALL\cite{ben2026training} & 0.556 / 0.807 & 0.536 / 0.767 & 0.582 / 0.859 & 0.558 / 0.811 \\

            \hline
        \end{tabular}
    \end{subtable}

    \vspace{0.3cm}

    \begin{subtable}{0.98\linewidth}
        \centering
        \caption{F1-score / ECE}
        \begin{tabular}{lcccc}
            \hline
            \textbf{Method} & Grok & VEO3.1 & Sora2 & Global \\
            \hline
            B-Free\cite{guillaro2025bias}& 0.278 / 0.116 & 0.223 / 0.133 & 0.326 / 0.104 & 0.295 / 0.109 \\
            C2P-CLIP\cite{tan2025c2p} & 0.666 / 0.015 & 0.664 / 0.014 & 0.665 / 0.014 & 0.853 / 0.015 \\
            ForgeLens\cite{chen2025forgelens}& 0.000 / 0.484 & 0.000 / 0.486 & 0.004 / 0.482 & 0.001 / 0.484 \\
            D3-Im\cite{yang2025d}& 0.054 / 0.463 & 0.075 / 0.446 &  0.058/ 0.464 & 0.063 / 0.457 \\
            C-Forensics\cite{park2025community}& 0.050 / 0.467 & 0.004 / 0.484 & 0.021 / 0.476 & 0.025 / 0.475 \\
            \hline
            D3-Vid\cite{zheng2025d3vid} & 0.679 / 0.291 & 0.684 / 0.285 & 0.660 / 0.301 & 0.847 / 0.122 \\
            L3DE\cite{chang2025l3de} & 0.139 / 0.592 & 0.110 / 0.595 & 0.224 / 0.550 & 0.186 / 0.722 \\
            NSG-VD\cite{zhang2025NSGVD} & 0.607 / 0.313 & 0.605 / 0.303 & 0.647 / 0.272 & 0.763 / 0.209 \\
            WaveRep\cite{corvi2025waverep} & 0.201 / 0.423 & 0.242 / 0.400 & 0.046 / 0.477 & 0.168 / 0.655 \\
            STALL\cite{ben2026training} & 0.407 / 0.201 & 0.397 / 0.209 & 0.422 / 0.200 & 0.408 / 0.203 \\

            \hline
        \end{tabular}
    \end{subtable}
    
    \vspace{0.3cm}
    
    \begin{subtable}{0.98\linewidth}
        \centering
        \caption{NLL}
        \begin{tabular}{lcccc}
            \hline
            \textbf{Method} & Grok & VEO3.1 & Sora2 & Global \\
            \hline
            B-Free\cite{guillaro2025bias} & 0.728 & 0.744 & 0.720 & 0.727 \\
            C2P-CLIP\cite{tan2025c2p} & 0.694 & 0.694& 0.694 &0.679  \\
            ForgeLens\cite{chen2025forgelens} & 2.252 & 2.302 & 2.216 & 2.257 \\
            D3-Im\cite{yang2025d} & 2.701 & 2.284 & 2.906 & 3.933 \\
            C-Forensics\cite{park2025community}& 3.033 & 3.423 & 2.981 & 4.715 \\
            \hline
            D3-Vid\cite{zheng2025d3vid} & 1.008 & 0.990 & 1.039 & 0.637 \\
            L3DE\cite{chang2025l3de} & 7.425 & 7.447 & 6.327 & 9.184 \\
            NSG-VD\cite{zhang2025NSGVD} & 1.050 & 1.045 & 0.957 & 0.786 \\
            WaveRep\cite{corvi2025waverep} & 1.750 & 1.492 & 2.659 & 2.936 \\
            STALL\cite{ben2026training} & 0.855 & 0.889 & 0.839 & 0.861 \\

            \hline
        \end{tabular}
    \end{subtable}

    \label{tab:Resolution}
\end{table}

\begin{table}[ht]
    \centering
    \caption{Results of image (top) and video (bottom) based methods on CCDF raw set with bitrates mirrored from real videos.}

    \begin{subtable}{0.98\linewidth}
        \centering
        \caption{AUC / AP}
        \begin{tabular}{lcccc}
            \hline
            \textbf{Method} & Grok & VEO3.1 & Sora2 & Global \\
            \hline
            B-Free\cite{guillaro2025bias} & 0.562 / 0.511 & 0.562 / 0.504& 0.591 / 0.513 & 0.572 / 0.747 \\
            C2P-CLIP\cite{tan2025c2p} & 0.503 / 0.520 & 0.605 / 0.613 & 0.532 / 0.556 & 0.547 / 0.789 \\
            ForgeLens\cite{chen2025forgelens} & 0.621 / 0.693 & 0.707 / 0.759 & 0.533 / 0.617 & 0.621 / 0.692 \\
            D3-Im\cite{yang2025d}& 0.527 / 0.562 & 0.592 / 0.606 & 0.520 / 0.559 & 0.546 / 0.797 \\
            C-Forensics\cite{park2025community}& 0.681 / 0.671 & 0.654 / 0.618 & 0.607 / 0.581 & 0.648 / 0.831 \\
            \hline
            D3-Vid\cite{zheng2025d3vid} & 0.548 / 0.525 & 0.557 / 0.542 & 0.601 / 0.622 & 0.569 / 0.793 \\
            L3DE\cite{chang2025l3de} & 0.392 / 0.441 & 0.447 / 0.490 & 0.460 / 0.513 & 0.433 / 0.733 \\
            NSG-VD\cite{zhang2025NSGVD} & 0.600 / 0.573 & 0.554 / 0.568 & 0.615 / 0.610 & 0.590 / 0.806 \\
            WaveRep\cite{corvi2025waverep} & 0.551 / 0.605 & 0.556 / 0.605 & 0.574 / 0.567 & 0.560 / 0.808 \\
            STALL\cite{ben2026training} & 0.587 / 0.573 & 0.543 / 0.538 & 0.634 / 0.624 & 0.588 / 0.579 \\

            \hline
        \end{tabular}
    \end{subtable}
    
    \vspace{0.3cm}
    
    \begin{subtable}{0.98\linewidth}
        \centering
        \caption{Accuracy / Recall}
        \begin{tabular}{lcccc}
            \hline
            \textbf{Method} & Grok & VEO3.1 & Sora2 & Global \\
            \hline
            B-Free\cite{guillaro2025bias} & 0.576 / 0.596 & 0.564 / 0.472 & 0.538 / 0.387 & 0.579 / 0.567 \\
            C2P-CLIP\cite{tan2025c2p} & 0.509 / 1.0 & 0.508 / 0.998 & 0.509 / 1.0 & 0.509 / 0.999 \\
            ForgeLens\cite{chen2025forgelens}& 0.514 / 0.030 & 0.512 / 0.026 & 0.505 / 0.013 & 0.510 / 0.023 \\
            D3-Im\cite{yang2025d}& 0.511 / 0.026 & 0.513 / 0.028 & 0.509 / 0.021 & 0.511 / 0.025 \\
            C-Forensics\cite{park2025community}& 0.501 / 0.002 & 0.501 / 0.002 & 0.500 / 0.000 & 0.500 / 0.001 \\
            \hline
            D3-Vid\cite{zheng2025d3vid} & 0.541 / 0.941 & 0.541 / 0.941 & 0.520 / 0.898 & 0.534 / 0.927 \\
            L3DE\cite{chang2025l3de} & 0.454 / 0.230 & 0.485 / 0.291 & 0.487 / 0.296 & 0.475 / 0.273 \\
            NSG-VD\cite{zhang2025NSGVD} & 0.567 / 0.857 & 0.533 / 0.787 & 0.556 / 0.835 & 0.552 / 0.826 \\
            WaveRep\cite{corvi2025waverep} & 0.530 / 0.065 & 0.524 / 0.052 & 0.500 / 0.004 & 0.518 / 0.041 \\
            STALL\cite{ben2026training} & 0.540 / 0.768 & 0.528 / 0.824 & 0.586 / 0.860 & 0.552 / 0.792 \\

            \hline
        \end{tabular}
    \end{subtable}

    \vspace{0.3cm}

    \begin{subtable}{0.98\linewidth}
        \centering
        \caption{F1-score / ECE}
        \begin{tabular}{lcccc}
            \hline
            \textbf{Method} & Grok & VEO3.1 & Sora2 & Global \\
            \hline
            B-Free\cite{guillaro2025bias}& 0.584 / 0.128 & 0.520 / 0.111 & 0.456 / 0.118 & 0.666 / 0.122 \\
            C2P-CLIP\cite{tan2025c2p} & 0.671 / 0.015 & 0.670 / 0.016 & 0.671 / 0.015 & 0.859 / 0.016 \\
            ForgeLens\cite{chen2025forgelens}& 0.059 / 0.455 & 0.051 / 0.444 & 0.026 / 0.469 & 0.045 / 0.456 \\
            D3-Im\cite{yang2025d}& 0.050 / 0.466 & 0.054 / 0.464 & 0.042 / 0.467& 0.049 / 0.465 \\
            C-Forensics\cite{park2025community}& 0.004 / 0.483 & 0.004 / 0.493 & 0.000 / 0.491 & 0.002 / 0.488 \\
            \hline
            D3-Vid\cite{zheng2025d3vid} & 0.672 / 0.293 & 0.672 / 0.291 & 0.651 / 0.319 & 0.838 / 0.125 \\
            L3DE\cite{chang2025l3de} & 0.297 / 0.513 & 0.361 / 0.487 & 0.366 / 0.487 & 0.395 / 0.601 \\
            NSG-VD\cite{zhang2025NSGVD} & 0.664 / 0.265 & 0.627 / 0.289 & 0.653 / 0.283 & 0.799 / 0.159 \\
            WaveRep\cite{corvi2025waverep} & 0.122 / 0.455 & 0.099 / 0.460 & 0.009 / 0.489 & 0.078 / 0.705 \\
            STALL\cite{ben2026training} & 0.404 / 0.210 & 0.398 / 0.209 & 0.429 / 0.196 & 0.410 / 0.205 \\

            \hline
        \end{tabular}
    \end{subtable}
    
    \vspace{0.3cm}
    
    \begin{subtable}{0.98\linewidth}
        \centering
        \caption{NLL}
        \begin{tabular}{lcccc}
            \hline
            \textbf{Method} & Grok & VEO3.1 & Sora2 & Global \\
            \hline
            B-Free\cite{guillaro2025bias} & 0.724 & 0.717 & 0.712 & 0.721 \\
            C2P-CLIP\cite{tan2025c2p} & 0.692 & 0.691 & 0.692 & 0.676 \\
            ForgeLens\cite{chen2025forgelens} & 1.842 & 1.685 & 2.004 & 1.843 \\
            D3-Im\cite{yang2025d} & 2.762 & 2.561 & 2.792 & 4.046 \\
            C-Forensics\cite{park2025community}& 3.476 & 3.699 & 3.928 & 5.548 \\
            \hline
            D3-Vid\cite{zheng2025d3vid} & 1.014 & 1.006 & 1.032 & 0.637 \\
            L3DE\cite{chang2025l3de} & 5.986 & 5.536 & 5.406 & 7.058 \\
            NSG-VD\cite{zhang2025NSGVD} & 0.925 & 1.000 & 0.936 & 0.674 \\
            WaveRep\cite{corvi2025waverep} & 2.777 & 2.781 & 2.836 & 4.189 \\
            STALL\cite{ben2026training} & 0.863  & 0.893 & 0.829  & 0.862  \\
            \hline
        \end{tabular}
    \end{subtable}

    \label{tab:Bitrates}
\end{table}

\begin{table}[ht]
    \centering
    \caption{Results of image (top) and video (bottom) based methods on CCDF raw set with 20s real videos (videos used to create prompts).}

    \begin{subtable}{0.98\linewidth}
        \centering
        \caption{AUC / AP}
        \begin{tabular}{lcccc}
            \hline
            \textbf{Method} & Grok & VEO3.1 & Sora2 & Global \\
            \hline
            B-Free\cite{guillaro2025bias} & 0.580 / 0.552 & 0.533 / 0.523 & 0.708 / 0.676 & 0.607 / 0.808 \\
            C2P-CLIP\cite{tan2025c2p}& 0.584 / 0.574 & 0.534 / 0.520 & 0.553 / 0.546 & 0.557 / 0.782 \\
            ForgeLens\cite{chen2025forgelens} & 0.744 / 0.764 & 0.658 / 0.675 & 0.574 / 0.634 & 0.658 / 0.693 \\
            D3-Im\cite{yang2025d}& 0.628 / 0.655 & 0.764 / 0.774 & 0.578 / 0.608 & 0.657 / 0.858 \\
            C-Forensics\cite{park2025community}& 0.773 / 0.766 & 0.771 / 0.776 & 0.764 / 0.766 & 0.770 / 0.903 \\
            \hline
            D3-Vid\cite{zheng2025d3vid} & 0.532 / 0.519 & 0.572 / 0.547 & 0.579 / 0.595 & 0.561 / 0.787 \\
            L3DE\cite{chang2025l3de} & 0.173 / 0.337 & 0.314 / 0.388 & 0.344 / 0.405 & 0.277 / 0.633 \\
            NSG-VD\cite{zhang2025NSGVD} & 0.674 / 0.622 & 0.708 / 0.663 & 0.740 / 0.721 & 0.707 / 0.858 \\
            WaveRep\cite{corvi2025waverep} & 0.971 / 0.971 & 0.926 / 0.940 & 0.718 / 0.749 & 0.872 / 0.957 \\
            STALL\cite{ben2026training} & 0.615 / 0.615 & 0.574 / 0.572 & 0.634 / 0.624 & 0.608 / 0.604 \\

            \hline
        \end{tabular}
    \end{subtable}
    
    \vspace{0.3cm}
    
    \begin{subtable}{0.98\linewidth}
        \centering
        \caption{Accuracy / Recall}
        \begin{tabular}{lcccc}
            \hline
            \textbf{Method} & Grok & VEO3.1 & Sora2 & Global \\
            \hline
            B-Free\cite{guillaro2025bias} & 0.558 / 0.483 & 0.528 / 0.309 & 0.618 / 0.459 & 0.580 / 0.425 \\
            C2P-CLIP\cite{tan2025c2p}& 0.513 / 0.997 & 0.511 / 0.993 & 0.513 / 0.998 & 0.512 / 0.996 \\
            ForgeLens\cite{chen2025forgelens}& 0.508 / 0.019 & 0.500 / 0.002 & 0.512 / 0.026 & 0.507 / 0.016 \\
            D3-Im\cite{yang2025d}& 0.521 / 0.046 & 0.539 / 0.080 & 0.520 / 0.043 & 0.527 / 0.056 \\
            C-Forensics\cite{park2025community}& 0.505 / 0.010 & 0.519 / 0.039 & 0.520 / 0.041 & 0.515 / 0.030 \\
            \hline
            D3-Vid\cite{zheng2025d3vid} & 0.536 / 0.950 & 0.543 / 0.965 & 0.515 / 0.909 & 0.531 / 0.941 \\
            L3DE\cite{chang2025l3de} & 0.380 / 0.037 & 0.409 / 0.093 & 0.423 / 0.122 & 0.404 / 0.084 \\
            NSG-VD\cite{zhang2025NSGVD} & 0.610 / 0.893 & 0.629 / 0.933 & 0.621 / 0.915 & 0.620 / 0.914 \\
            WaveRep\cite{corvi2025waverep} & 0.800 / 0.615 & 0.788 / 0.591 & 0.567 / 0.150 & 0.719 / 0.452 \\
            STALL\cite{ben2026training} & 0.830 / 0.264 & 0.538 / 0.812 & 0.564 / 0.864 & 0.550 / 0.836 \\

            \hline
        \end{tabular}
    \end{subtable}

    \vspace{0.3cm}

    \begin{subtable}{0.98\linewidth}
        \centering
        \caption{F1-score / ECE}
        \begin{tabular}{lcccc}
            \hline
            \textbf{Method} & Grok & VEO3.1 & Sora2 & Global \\
            \hline
            B-Free\cite{guillaro2025bias}& 0.522 / 0.125 & 0.396 / 0.107 & 0.546 / 0.117 & 0.561 / 0.109\\
            C2P-CLIP\cite{tan2025c2p} & 0.672 / 0.015 & 0.670 / 0.14 & 0.672 / 0.015 & 0.859 / 0.015 \\
            ForgeLens\cite{chen2025forgelens}& 0.038 / 0.450 & 0.004 / 0.469 & 0.050 / 0.462 & 0.031 / 0.461 \\
            D3-Im\cite{yang2025d}& 0.087 / 0.442 & 0.148 / 0.411 & 0.083 / 0.448 & 0.106 / 0.434 \\
            C-Forensics\cite{park2025community}& 0.021 / 0.469 & 0.075 / 0.458 & 0.079 / 0.463 & 0.059 / 0.463 \\
            \hline
            D3-Vid\cite{zheng2025d3vid} & 0.672 / 0.300 & 0.679 / 0.308 & 0.652 / 0.329 & 0.843 / 0.126 \\
            L3DE\cite{chang2025l3de} & 0.056 / 0.605 & 0.137 / 0.565 & 0.174 / 0.547 & 0.143 / 0.734 \\
            NSG-VD\cite{zhang2025NSGVD} & 0.696 / 0.240 & 0.716 / 0.242 & 0.707 / 0.258 & 0.855 / 0.086 \\
            WaveRep\cite{corvi2025waverep} & 0.755 / 0.177 & 0.736 / 0.199 & 0.258 / 0.402 & 0.621 / 0.401 \\
            STALL\cite{ben2026training} & 0.368 / 0.218 & 0.363 / 0.216 & 0.377 / 0.219 & 0.369 / 0.218 \\

            \hline
        \end{tabular}
    \end{subtable}
    
    \vspace{0.3cm}
    
    \begin{subtable}{0.98\linewidth}
        \centering
        \caption{NLL}
        \begin{tabular}{lcccc}
            \hline
            \textbf{Method} & Grok & VEO3.1 & Sora2 & Global \\
            \hline
            B-Free\cite{guillaro2025bias} & 0.715 & 0.712 & 0.688 & 0.706 \\
            C2P-CLIP\cite{tan2025c2p} & 0.692 & 0.692 & 0.692 & 0.677 \\
            ForgeLens\cite{chen2025forgelens} & 1.681 & 1.901 & 1.924 & 1.835 \\
            D3-Im\cite{yang2025d} & 2.288 & 1.739 & 2.499 & 3.247 \\
            C-Forensics\cite{park2025community}& 2.784 & 2.274 & 2.802 & 4.162 \\
            \hline
            D3-Vid\cite{zheng2025d3vid} & 1.040 & 1.026 & 1.056 & 0.640 \\
            L3DE\cite{chang2025l3de} & 8.192 & 6.955 & 6.650 & 9.825 \\
            NSG-VD\cite{zhang2025NSGVD} & 0.848 & 0.819 & 0.838 & 0.541 \\
            WaveRep\cite{corvi2025waverep} & 0.478 & 0.716 & 1.959 & 1.559 \\
            STALL\cite{ben2026training} & 0.835 & 0.863 & 0.827 & 0.842 \\

            \hline
        \end{tabular}
    \end{subtable}

    \label{tab:duration}
\end{table}

\FloatBarrier


\newpage

\end{document}